\documentclass[runningheads]{llncs}

\usepackage{eccv}

\usepackage{eccvabbrv}

\usepackage{graphicx}
\usepackage{booktabs}

\usepackage[accsupp]{axessibility}  %

\usepackage{hyperref}
\hypersetup{
    colorlinks=true,
    citecolor=RoyalPurple,
}
\usepackage{orcidlink}

\usepackage{lipsum}
\usepackage{amsmath}
\usepackage{multirow}
\usepackage[normalem]{ulem}
\useunder{\uline}{\ul}{}
\usepackage{subcaption}
\usepackage{adjustbox}
\usepackage{colortbl}
\usepackage{animate}
\usepackage{amssymb}
\usepackage{bm}
\usepackage{wrapfig}

\definecolor{darkergreen}{RGB}{0, 0, 0}
\newcommand\greenp[1]{\textcolor{darkergreen}{(#1)}}
\definecolor{red2}{RGB}{252, 54, 65}

\newcommand\greenpscript[1]{\scriptsize\greenp{#1}}

\newcommand{\ours}{{Model Stock}\xspace}

\newcommand{\ourss}{{Model Stock$^\star$}\xspace}

\definecolor{bleudefrance}{rgb}{0.19, 0.55, 0.91}
\definecolor{Gray}{gray}{0.85}

\renewcommand{\lipsum}[1][1-150]{} %

\begin{document}

\title{Model Stock: All we need is just a few fine-tuned models}

\titlerunning{Model Stock}

\author{
Dong-Hwan Jang\thanks{Work done during an internship at NAVER AI Lab.} \quad 
Sangdoo Yun \quad 
Dongyoon Han 
}

\authorrunning{Jang et al.}

\institute{
NAVER AI Lab 
}

\maketitle
\begin{abstract}
This paper introduces an efficient fine-tuning method for large pre-trained models, offering strong in-distribution (ID) and out-of-distribution (OOD) performance. Breaking away from traditional practices that need a multitude of fine-tuned models for averaging, our approach employs significantly fewer models to achieve final weights yet yield superior accuracy. Drawing from key insights in the weight space of fine-tuned weights, we uncover a strong link between the performance and proximity to the center of weight space. Based on this, we introduce a method that approximates a center-close weight using only two fine-tuned models, applicable during or after training. Our innovative layer-wise weight averaging technique surpasses state-of-the-art model methods such as Model Soup, utilizing only two fine-tuned models. This strategy can be aptly coined \textit{\ours}, highlighting its reliance on selecting a minimal number of models to draw a more optimized-averaged model. We demonstrate the efficacy of \ours with fine-tuned models based upon pre-trained CLIP architectures, achieving remarkable performance on both ID and OOD tasks on the standard benchmarks, all while barely bringing extra computational demands. Our code and pre-trained models are available at \url{https://github.com/naver-ai/model-stock}. 
\end{abstract}    
\section{Introduction}
\label{sec:intro}

\begin{figure}[t]
    \centering
    \includegraphics[width=0.7\linewidth]{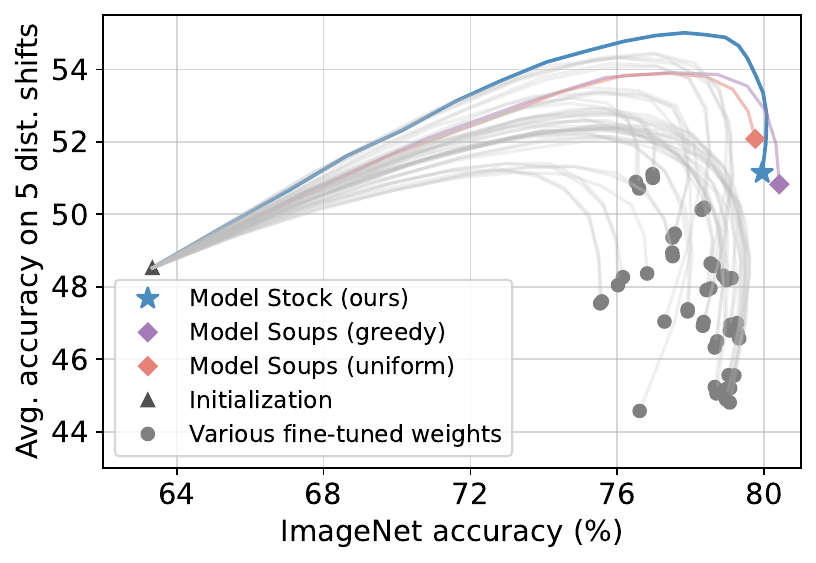}
    \vspace{-1em}
    \caption{{\textbf{\ours vs. Model Soup.} 
    \ours consistently enjoys improved accuracy on ImageNet (x-axis) and distribution shift benchmarks (y-axis) against individual fine-tuned models (gray circles). We plot WiSE-FT~\cite{wiseft} curves to each fine-tuned model, highlighting \ours's better performance on distribution shifts compared to Model Soup~\cite{modelsoup}.
    Note that \ours has much smaller computational costs than Model Soups (24$\times$ smaller), \ie, \ours requires two fine-tuning procedures, whereas Model Soups are leveraging 48 various fine-tuned models in this experiment.
    }}
    \label{fig:teaser}  
    \vspace{-1.5em}
\end{figure}

Pre-train/fine-tune paradigm~\cite{lpft,wiseft,modelsoup,li2022trainable,tian2023fast} has proven to be a strong framework for training models to reach state-of-the-art performance. This approach, especially pivotal in fine-tuning pre-trained models, involves models acquiring general knowledge during pre-training and task-specific knowledge during fine-tuning. How we perform a fine-tuning stage is crucial, affecting task performance and robustness against distribution shifts.

Recent advancements, notably Model Soup~\cite{modelsoup}, which merges weights from multiple fine-tuned models trained under different training setups, have shown impressive performance without increasing inference costs. This method is believed to be effective because these models often reside in the same loss basin, and their merging results in a lower and flat loss basin. However, Model Soup's requirement for multiple fine-tuned models (more than dozens) raises concerns about efficiency and practicality in general scenarios where the models need to be prepared from scratch. Thus, our question is: Is there an efficient way to achieve an effective merged weight from very few fine-tuned models?

We initially explore the dynamics of fine-tuned weights under the simplest scenario: varying the random seeds while maintaining the other training setups. It reveals that the fine-tuned weights with different random seeds reside on a very thin shell layer-wise during and after training.
We then delve into the impact of a model-soup-like weight averaging approach. Our findings show that the closer proximity of the averaged weights correlates with improved In-Distribution (ID) and Out-Of-Distribution (OOD) performance.

Building upon the findings, we propose a novel approach of fine-tuning method coined \textbf{\ours}, analogous to chicken stock in cooking, distinguishing it from what Model Soup intended. Now, the answer to our question is indeed affirmative: \ours approximates the merged weight using just a few fine-tuned models, leveraging the weight space's geometric properties and a pre-trained model's anchoring effect. This strategy offers a more computationally efficient alternative to the labor-intensive averaging of fine-tuned models, streamlining the process while enhancing model performance. 
Fig.~\ref{fig:teaser} illustrates our brief comparison of \ours vs. Model Soup~\cite{modelsoup}. 
We reproduce the Model Soup experiments\footnote{We follow the standard grid hyper-parameter sweep\cite{modelsoup}, fine-tuning for 10 epochs. 
Our results align with those presented in Fig. D.1 (right) of the original paper \cite{modelsoup}.} based on the CLIP ViT-B/32 initialization by fine-tuning 48 models with various hyper-parameters, which is defined as \textit{zero-shot} initialization setting.
Fig.~\ref{fig:teaser} shows that \ours outperforms Model Soups with much smaller computational costs.

Our comprehensive experiments demonstrate the effectiveness of \ours. We achieve performance comparable to, or even surpassing, that of the more resource-intensive methods such as Model Soup~\cite{modelsoup}, using only a fraction of the models. Specifically, our method achieves 87.8\% ImageNet top-1 accuracy (ID) and averaged 74.9\% in five distribution shift benchmarks (OOD) on ViT-L/14 fairly compared with the prior arts using the CLIP pre-trained weight. 
We believe that our study not only underscores \ours's practicability but also opens new directions in the pre-train/fine-tune paradigm for superior performance across various tasks.

\section{Analyzing Fine-tuned Weights}
\label{sec:intuition}

Our study is driven by two fundamental findings related to the performance and robustness of fine-tuned models. 
The first one is that model weights are fine-tuned on different random seeds\footnote{Random seed influences training randomness, such as training data shuffling and data augmentation parameters.} lie on \textit{thin shell} in weight space layer-wise. %
The second posits that closer proximity to the center of this thin shell is beneficial for improving performance across the ImageNet and distribution shift benchmarks. 
The substantiation and implications of these observations are discussed in the subsequent sections.

\vspace{-1em}
\subsection{Geometric Properties Between Weights} 
\label{subsec:geometric_prop_bw_weights}
\subsubsection{Angle and norm of weights.}
We begin by examining the intrinsic properties of the weights in fine-tuned models. We define the weight vector of the fine-tuned model at $k$-th layer as $\mathbf{w}^{(k)} \in \mathbb{R}^{n^{(k)}}$ where $n^{(k)}$ is the number of weight parameters at $k$-th layer, and the origin $\mathbf{0}$ as the pre-trained model weight $\mathbf{w}_0^{(k)}$ at $k$-th layer.
Then the angle $\theta^{(k)}$ between two weight $\mathbf{w}^{(k)}_1$ and $\mathbf{w}^{(k)}_2$ is defined by:
    $\theta^{(k)} = \arccos\left( \frac{\mathbf{w}^{(k)}_1 \cdot \mathbf{w}^{(k)}_2}{\| \mathbf{w}^{(k)}_1 \| \| \mathbf{w}^{(k)}_2 \|} \right),$
where the Euclidean $l_2$-norm $\|\mathbf{w}^{(k)}\|$ of a $n^{(k)}$ defined as 
$\|\mathbf{w}^{(k)}\|  = \sqrt{\sum_{i=1}^{n^{(k)}} {w_i^{(k)}}^2}.$ 
Angle and norm will provide a geometric view of the weights at $k$-th layer between fine-tuned models.

\begin{figure}[t]
\centering
    \includegraphics[width=1\linewidth]{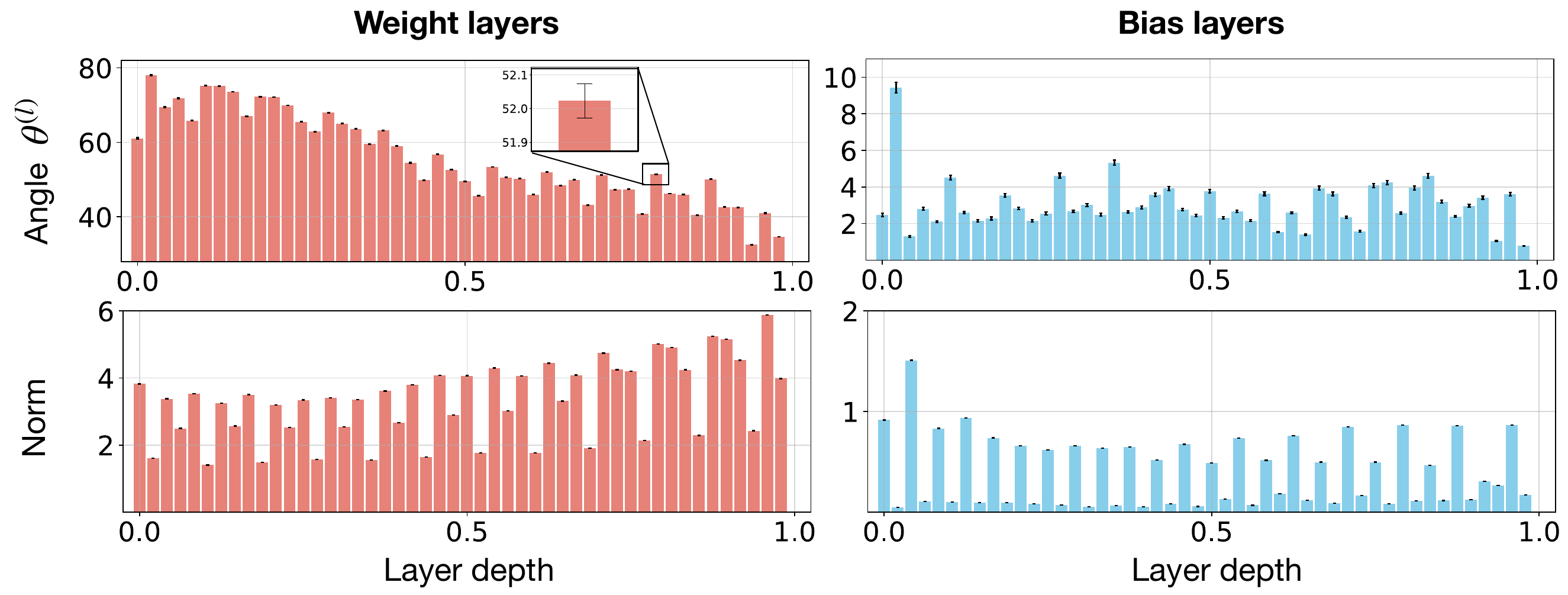} %
    \vspace{-1.5em}
    \caption{
    \textbf{Layer-wise angle and norm of fine-tuned models.} 
    We measure the angle $\theta^{(k)}$ (degree) and norm $\frac{\|\mathbf{w}^{(k)}\|}{\sqrt{n^{(k)}}}$ for 50 distinct weights, fine-tuned under different random seeds.    
    We separately visualize weight layers (red bars) and bias layers (blue bars), where bias layers have much smaller angles.
    We plot the mean angle and norm values with standard deviation (black error bar).
    The results show that any two fine-tuned weights have \textit{layer-wise consistent} angle and norm with extremely low standard deviation.     
    }
    \label{fig:angle}
    \vspace{-1em}
\end{figure}

\vspace{-1em}
\subsubsection{Observation 1: Angle and norm consistency among fine-tuned weights.} 
We investigate the weight space of models fine-tuned on ImageNet from a pre-trained model with various random seeds.
Our first observation is that both angle $\theta^{(k)}$ between two different models and norm $\|\mathbf{w}^{(k)}\|$ of a weight exhibit \textit{consistent} values with very low standard deviations, as shown in Fig.~\ref{fig:angle}.
This consistency can be mathematically represented as follows: For all \( i \) and \( j \in [1, N] \) when the number of fine-tuned weights, \( N \), is sufficiently large, the following holds:
\begin{equation}
\mathbf{w}_i^{(k)} \cdot \mathbf{w}_j^{(k)} = 
\begin{cases} 
\left(l^{(k)}\right)^2 & \text{if } i = j, \\
\left(l^{(k)}\right)^2 \cos\theta^{(k)} & \text{otherwise},
\end{cases}
\label{eq:obs}
\end{equation}
where \( l^{(k)} \) and \( \theta^{(k)} \) are constants that describe the magnitude and angle between weights at $k$-th layer, respectively. Henceforth, to simplify notation, we will omit the superscript \((k)\) indicating the layer index.

Interestingly, these consistencies in angle and norm are observed 1) across diverse setups and 2) both during and after training. 
Fig.~\ref{fig:angle} shows this consistency over 50 fine-tuned CLIP ViT-B/32\footnote{\url{https://github.com/openai/CLIP}}~\cite{clip} 
models. %
It illustrates that the \textit{layer-wise} norm and angle of these models exhibit almost constant values with extremely minimal error. 
While this figure depicts a specific model (\ie, CLIP ViT-B/32), we establish that such regularity is not confined to a single model or setting but is consistent across various CLIP fine-tuning scenarios. 
We conjecture this holds irrespective of networks (ViT~\cite{dosovitskiy2020vit}, Hybrid-ViT~\cite{dosovitskiy2020vit}, ResNet~\cite{resnet}, ConvNext~\cite{convnext}), optimizers (SGD, AdamW~\cite{adamw}), data augmentations (RRC~\cite{inception}, RandAug~\cite{randaug}), datasets (CIFAR~\cite{cifar}, ImageNet~\cite{imagenet}), or initialization of the classifier (zero-shot, LP-FT~\cite{lpft}). 
Remarkably, this regularity also holds for fine-tuned weights at each step during training as well as after training.
A comprehensive analysis supporting these findings is presented in the Appendix~\ref{suppl:consistency}.

Based on the observation, we presume the distribution of the fine-tuned weights. The center of the fine-tuned weights is defined as \(\boldsymbol{\mu}\)=\(\lim_{N \rightarrow \infty} \frac{1}{N} \sum_{i=1}^N \mathbf{w}_i\). 
We then deduce the following properties among fine-tuned weights: (i) \( \|\mathbf{w}_i - \boldsymbol{\mu} \| = \text{constant} \), indicating a \textit{thin shell} distribution; (ii) \( (\mathbf{w}_0-\boldsymbol{\mu}) \perp (\mathbf{w}_i-\boldsymbol{\mu}) \); and (iii) \( (\mathbf{w}_i-\boldsymbol{\mu}) \perp (\mathbf{w}_j-\boldsymbol{\mu}) \) for all \( i, j \in [1, N] \). These properties are depicted in Fig.~\ref{fig:gaussian} for better understanding. The detailed proof is in the Appendix~\ref{supp:proof_geometric}.

\begin{table}[t]
\centering
\small
\caption{\textbf{Distance from the center (\ie, $\|\mathbf{w-\boldsymbol{\mu}}\|$) vs. performance}. We report the ImageNet and distribution shift performance with the distance from the center of weights $\mu$ for fine-tuned and averaged models. We observe 1) both models consistently maintain a nearly constant distance from $\mu$ with remarkably small standard deviation; 2) averaging more models approaches $\mu$, boosting ID and OOD performance. }
\label{tab:avr_perform}
\vspace{-1em}
\tabcolsep=0.75em
\resizebox{.6\linewidth}{!}{
\begin{tabular}{lccc}
\toprule
   & $\|\mathbf{w}-\boldsymbol{\mu}\|$ & ImageNet  & Avg. shifts   \\ 
   \midrule
Fine-tuned & 13.133$\pm$.004& 79.72 & 46.37          \\
$\mathbf{w}_\text{avr}^{(2)}$     & 9.192$\pm$.003   & 80.24 & 47.76    \\
$\mathbf{w}_\text{avr}^{(3)}$  & 7.439$\pm$.025   & 80.35 & 48.18       \\
$\mathbf{w}_\text{avr}^{(5)}$    & 5.633$\pm$.014  & 80.47 & 48.53     \\
$\mathbf{w}_{\text{avr}}^{(50)}\simeq\boldsymbol{\mu}$     & $\sim$0 & 80.59 & 48.85     \\
\bottomrule
\end{tabular}
}
\vspace{-1em}
\end{table}

\subsection{Center of Weights and Performance}
\label{subsec:weight_center}

We proceed to explore the relationship between the proximity to the center of fine-tuned weights and their performance on ID and OOD datasets.
Given that computing the exact center is infeasible, we approximate it by averaging differently seeded 50 fine-tuned weights, using it as a pseudo-center (\ie $\boldsymbol{\mu}{\simeq}\mathbf{w}_{\text{avr}}^{(50)}$).

\subsubsection{Observation 2: Distance from the center of weights and performance.}  
Table~\ref{tab:avr_perform} offers quantitative observations about the fine-tuned weights and their performance using CLIP ViT-B/32. 
The results include distances from the weight center ($\boldsymbol{\mu}$) of fine-tuned models and their averaged counterparts with their ID and OOD performances. 
Going closer to the center by averaging the weights leads to improving both performances.
Interestingly, the standard deviation of the distances is less than 0.1\% of the mean distance, suggesting highly consistent fine-tuned weight distances from the center across different weights. 
This suggests that fine-tuned weights occupy a \textit{thin shell} as discussed in \S\ref{subsec:geometric_prop_bw_weights}.

\begin{figure}[t]
    \centering
    \includegraphics[width=.8\textwidth]{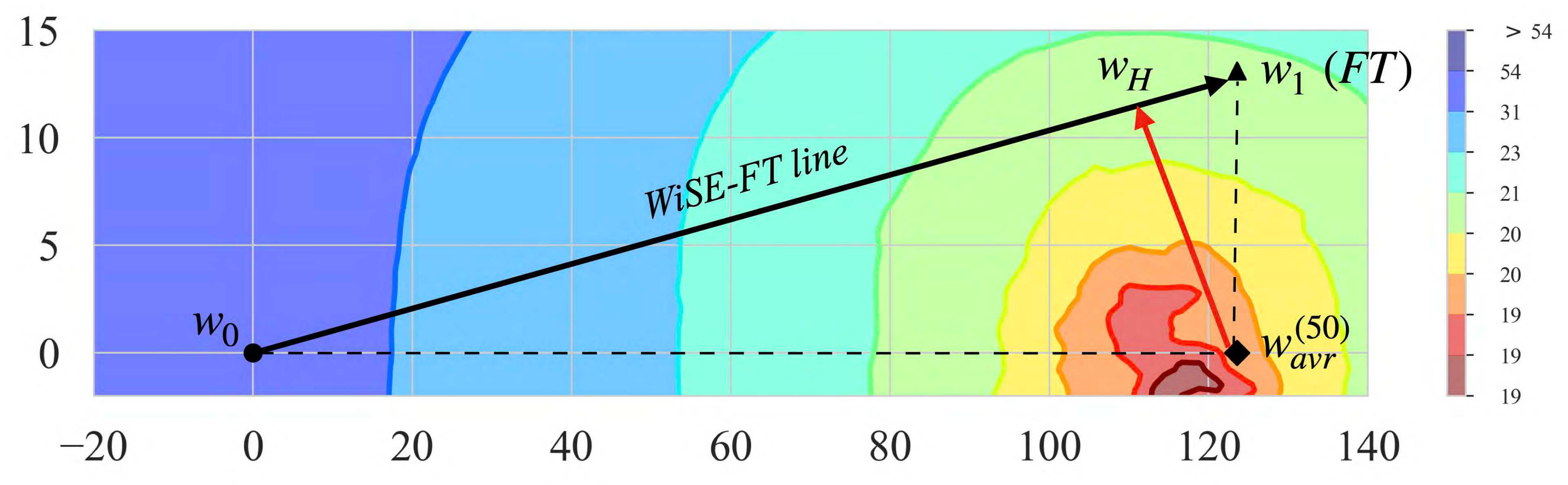}
    \vspace{-1em}
    \caption{\textbf{Test error landscape.} %
    We visualize a test error landscape with three weight points: a pre-trained model ($\mathbf{w_0}$), a fine-tuned model ($\mathbf{w_1}$), and the averaged weights of 50 fine-tuned models ($\mathbf{w}_\text{avr}^{50}$). 
    $\mathbf{w}_\text{avr}^{50}$ locates near the lowest error basin. 
    A better solution ($\mathbf{w}_H$) with lower error than $\mathbf{w_1}$ can be easily found utilizing those three weights (\S\ref{sec:method}). 
    } 
    \label{fig:landscape}
    \vspace{-1em}
\end{figure}

\vspace{-1em}
\subsubsection{Observation 3: Fine-tuned weights occupy local minima edges in the test error landscape.} 
We present an additional observation regarding the test error landscape, which relates to the performance around the center of weights. Fig.~\ref{fig:landscape} depicts the test error landscape on the ImageNet test dataset within a two-dimensional plane. This plane includes a pre-trained model's weight ($\mathbf{w}_0$), a single fine-tuned model ($\mathbf{w}_1$), and the pseudo-center ($\mathbf{w}_\text{avr}^{(50)}$). 
This landscape reveals that a fine-tuned model typically occupies the boundary of test error regions. 
On the other hand, centered near pseudo-center ($\mathbf{w}_\text{avr}^{(50)}$), the test error is the lowest and gets higher as the weights get far from the center.
Interestingly, along the line segment $\overline{\mathbf{w}_0\mathbf{w}_1}$ (\ie, a WiSE-FT curve~\cite{wiseft}), the fine-tuned weight $\mathbf{w}_1$ is neither the point closest to the pseudo-center nor the one with the lowest test error.
We will connect this observation in \S\ref{sec:method} 
to find $\mathbf{w}_H$, the weight on the line closest to the center.

\subsubsection{Observation 4: Randomly perturbed weights nearing the center also merit high performance.} 
\begin{figure}[t]
    \centering
    \includegraphics[width=0.6\textwidth]{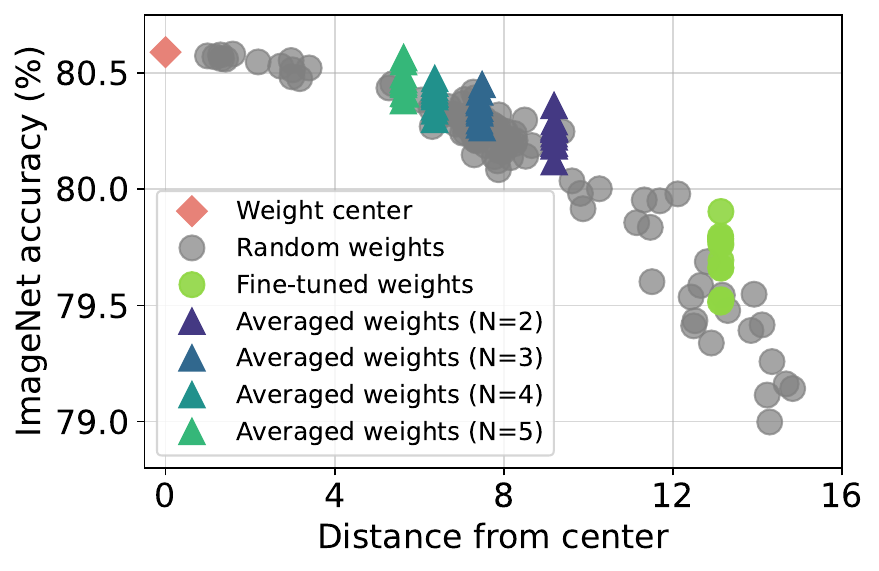}
    \vspace{-.5em}
    \caption{\textbf{Weights closer to the center enjoy higher performances.} 
    We show the ImageNet accuracy of weights with their distance from the center ($\boldsymbol{\mu}$), which is approximated with $\mathbf{w}_\text{avr}^{(50)}$.    
    Averaged weights are closer to the center than individual weights, and their accuracy increases as the averaging number of models ($N$) increases. 
    Gray circles are the weights randomly sampled from the Gaussian distribution centered at $\mathbf{w}_\text{avr}^{(50)}$.
    Even the random weights also achieve higher performance as they reach the center. 
    The results indicate the critical role of proximity to the center on performance. 
    }
    \label{fig:random_noise_weight}    
    \vspace{-1em}
\end{figure}

To further investigate the impact of proximity to the center on performance, we conduct a toy experiment to measure ImageNet performance using random weights generated by adding layer-wise noise to the weight at the center. 
The standard deviation of the noise for each layer is adjusted to align with the distribution of fine-tuned weights. 
Fig.~\ref{fig:random_noise_weight} presents a scatter plot correlating the distance of model weights from the distribution's center with corresponding ImageNet accuracies. 
Remarkably, randomly sampled weights demonstrate performance comparable to fine-tuned and averaged models, highlighting the importance of center proximity.

Finally, the above observations naturally raise a question: Why do fine-tuned weights through optimization not reach the center, staying constantly close instead? %
Previous studies~\cite{izmailov2018averaging, cha2021swad} tell us that optimization steps might struggle to guide fine-tuned weights to the center of the weight distribution due to the many stationary points in the loss surface. %
Alternatively, averaging independently fine-tuned models is a unique solution but both laborious and resource-intensive. %
It appears that there are no better alternatives for getting closer to the center, as optimization proves ineffective near these flat local minima. Could there be a faster approach to reaching the center? This question will be addressed in \S\ref{sec:method}.

\begin{figure}[t]
\centering
    \includegraphics[width=0.7\linewidth]{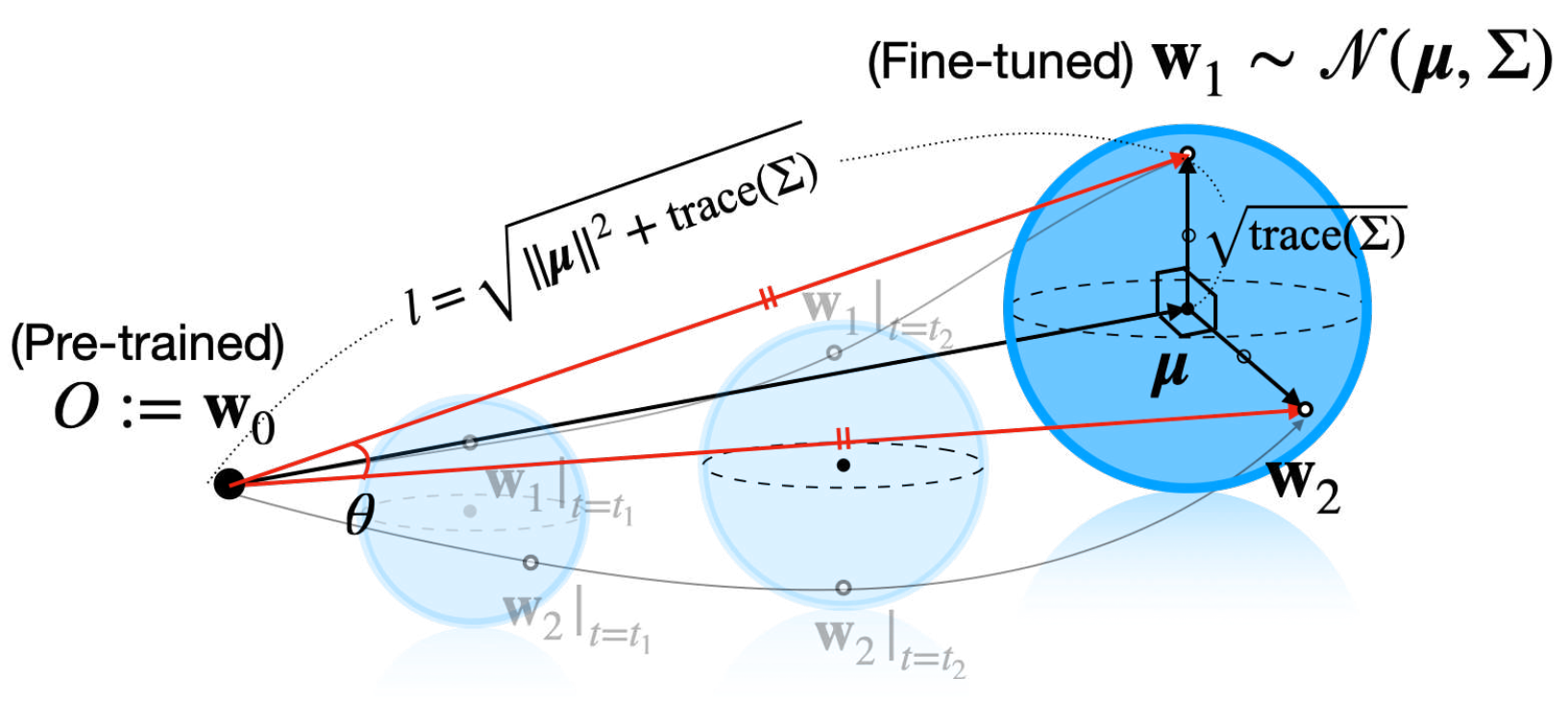}
    \vspace{-.5em}
    \caption{\textbf{Comprehensive illustration of the geometric dynamics of fine-tuned weights.} 
    This figure illustrates the behavior of fine-tuned weights that lie on a thin shell, supporting our Gaussian distribution hypothesis.
    Each sphere represents a thin shell that fine-tuned models lie on at each training step, with $\mathbf{w}_0$ denoting the pre-trained model. The curved lines trace the fine-tuning trajectory from $\mathbf{w}_0$, while the red vectors indicate that fine-tuned models at each time step are equidistant from $\mathbf{w}_0$.}
\label{fig:gaussian}
    \vspace{-1em}
\end{figure}

\subsection{Our Hypothesis}
The observed geometric patterns in weight distributions closely align with mathematical properties of Gaussian distributions, represented as $\mathcal{N}(\boldsymbol{\mu}, \Sigma)$. 
Therefore, a plausible reason for the particular geometric pattern of fine-tuned weights could be the influence of Gaussian noise within the weight space.
In high-dimensional spaces, vectors sampled from such distributions tend to have nearly identical norms, specifically $\sqrt{|\boldsymbol{\mu}|^2 + \text{trace}(\Sigma)}$ and consistent in-between angles, due to the \textit{concentration of measure phenomenon}~\cite{ledoux2001concentration}.
The likelihood of the squared norm significantly deviating from this expected value is exponentially negligible in high-dimensional spaces, like a weight space. 

In other words, the vectors sampled from high-dimensional Gaussian distribution lie on a very thin shell with the radius $\approx\sqrt{\text{trace}(\Sigma)}$ around the center $\boldsymbol{\mu}$. 
Consequently, we hypothesize that fine-tuned weights follow a Gaussian distribution in a layer-wise manner.
While not necessary for our observation, this hypothesis provides a sufficient condition for understanding the geometric dynamics of fine-tuned weights. 
For example, it aids in the intuitive understanding of the fact that the distance of $\mathbf{w}_{\text{avr}}^{(N)}$ from the weight center is proportional to $1/\sqrt{N}$, signifying the reduction of variance.
Fig.~\ref{fig:gaussian} comprehensively illustrates the observations and our hypothesis discussed in \S\ref{sec:intuition}.

\section{Method}
\label{sec:method}

This section introduces our method, \ours, a cost-efficient weight merging method.
As discussed in \S\ref{subsec:weight_center}, getting closer to the center of weights $\boldsymbol{\mu}$ induces improved model performance. 
A straightforward method to approximate $\boldsymbol{\mu}$ is averaging multiple model weights, which can be computationally expensive. 
We propose an efficient alternative method by leveraging \textit{the pre-trained model weights}, an aspect previously neglected by existing weight-merging methods.
A pre-trained model usually possesses general knowledge and shows robust and reliable performance in out-of-distribution (OOD) cases~\cite{clip}. 
Therefore, a pre-trained model can become a \textit{robust anchor point}. As shown in Fig.~\ref{fig:landscape}, we could readily identify a weight ($\mathbf{w}_H$) that is closer to the center---and thus better---by interpolating with the anchor. %
Building on this concept, we propose a method to approximate the center of weights more accurately with only a few fine-tuned weights. 
Again, for readability, we omit the layer notation $(k)$, but the subsequent method is applied layer-wise.
\begin{figure}[t]
    \centering
    \centering
    \includegraphics[width=0.75\textwidth, keepaspectratio]{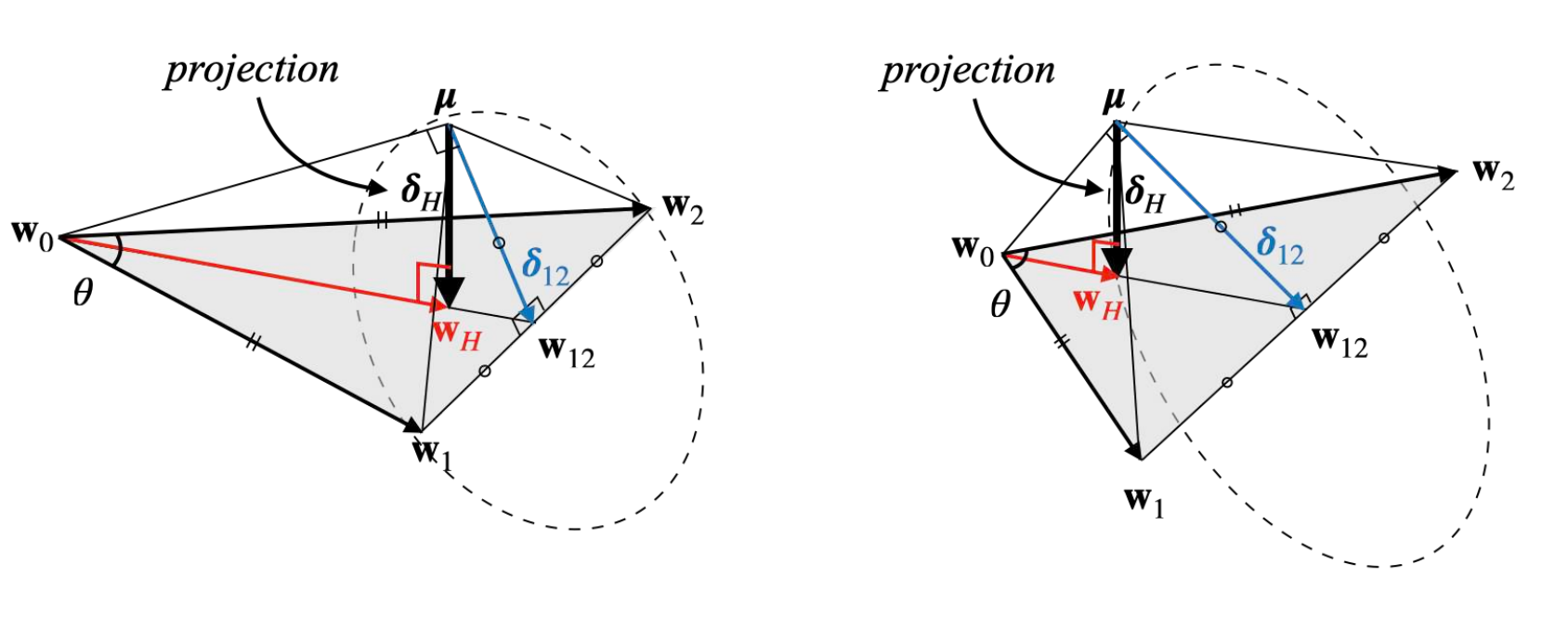}
    \label{fig:method_two}
    \vspace{-2em}
    \caption{
    \textbf{Schematic concept of our \ours.} We present two scenarios with a small angle (left) and a large angle (right).     
    Given a pre-trained weight (\(\mathbf{w}_0\)) and two fine-tuned weights (\(\mathbf{w}_1\), and \(\mathbf{w}_2\)), we visualize a gray triangle representing the span of three weights. We consider this triangle area our search space spanned by three weights.      
    We aim to find the best weight point on the triangle nearest the ideal center $\boldsymbol{\mu}$.     
    We find that the perpendicular foot \(\mathbf{w}_H\) from point $\boldsymbol{\mu}$ to the plane is the nearest point, which can be specified solely using the angle between the fine-tuned models, even without knowing the exact position of the center \(\boldsymbol{\mu}\).
    We utilize \(\mathbf{w}_H\) as the merged weight of \ours.
    Intuitively, when the angle $\theta$ is large (\ie, two weights are diverse), as in the right figure, \(\mathbf{w}_H\) will rely more on the pre-trained weight(\(\mathbf{w}_0\)), vice versa. 
    }
    \label{fig:method_2_and_n}
    \vspace{-1em}
\end{figure}

\noindent\textbf{On two fine-tuned models.}
We observed in \S\ref{sec:intuition} that two fine-tuned models with different random seeds have almost constant norms and the angle between them. 
Based on this, we define a plane connecting the pre-trained model and two fine-tuned models as shown in Fig.~\ref{fig:method_2_and_n} (depicted as a gray triangular area). 
Any weight vector on this plane can be expressed as a linear combination of the pre-trained model and two fine-tuned models.
Our goal is to find the weight closest to the center of fine-tuned weights on this plane, which is the perpendicular foot ($\mathbf{w}_H$) from the center of distribution ($\boldsymbol{\mu}$) to the plane.

Even without knowing the exact position of the center $\boldsymbol{\mu}$, \(\mathbf{w}_H\) can be specified solely using the angle between the fine-tuned models with the following two conditions that $\boldsymbol{\mu}$ must satisfy.
First, as mentioned in \S\ref{subsec:geometric_prop_bw_weights}, \( (\mathbf{w}_1-\boldsymbol{\mu}) \perp (\mathbf{w}_2-\boldsymbol{\mu}) \) and \( \|\mathbf{w}_i - \boldsymbol{\mu} \| = \|\mathbf{w}_2 - \boldsymbol{\mu} \| \) hold, implying that $\triangle \mathbf{w}_1 \boldsymbol{\mu}\mathbf{w}_2$ forms an isosceles right triangle.
In Fig.~\ref{fig:method_2_and_n}, $\boldsymbol{\mu}$ should lie on the dotted hyper-circle.
Another condition is that the condition \( (\mathbf{w}_0-\boldsymbol{\mu}) \perp (\mathbf{w}_{12}-\boldsymbol{\mu}) \) must be satisfied, where $\mathbf{w}_{12}=\frac{\mathbf{w}_1 + \mathbf{w}_2}{2}$.
This condition arises from the second property in \S\ref{subsec:geometric_prop_bw_weights}, where $(\mathbf{w}_0-\boldsymbol{\mu}) \perp (\mathbf{w}_i-\boldsymbol{\mu})$.
Combining these two conditions, $\boldsymbol{\mu}$ is at the point where the line starting from $\mathbf{w}_0$ is tangent to the hyper-circle.
We precisely determine the position of $\boldsymbol{\mu}$ within a 3D volume slice that encompasses both $\boldsymbol{\mu}$ and $\triangle\mathbf{w}_0 \mathbf{w}_1 \mathbf{w}_2$.
Consequently, we can find the position of the closest weight $\mathbf{w}_H$ to the distribution's center on the plane using straightforward geometric principles.
The position of the perpendicular foot is determined as follows:
\begin{equation}
    \mathbf{w}_H = \frac{2\cos\theta}{1+\cos\theta} \cdot \mathbf{w}_{12} + \left(1-\frac{2\cos\theta}{1+\cos\theta}\right) \cdot \mathbf{w}_0.
    \label{eq:two_inter_ratio}
\end{equation}
For more detailed proof, please refer to the Appendix~\ref{suppl:n_proof}. 
Note that the interpolation ratio $t=\frac{2\cos\theta}{1+\cos\theta}$ is solely determined by the angle $\theta$ between two fine-tuned models. 
Crucially, unlike previous methods, determining $t$ does not require extra training~\cite{modelsoup, li2022trainable, tian2023trainable, tian2023fast} or heuristic hyper-parameter settings~\cite{gouk2020distance, modelsoup}, thereby simplifying the process and enhancing its accessibility and efficiency.

As the $\theta$ decreases, the pre-trained model is less utilized for merging, as shown in Fig~\ref{fig:method_2_and_n} (left). 
Coupled with the observation in Fig.~\ref{fig:angle}, this indicates that bias layers rely less on pre-trained models and focus more on fine-tuned models, whereas weight layers depend more on pre-trained models. This observation extends the findings of previous works such as BitFit~\cite{zaken2021bitfit} and LP-FT~\cite{lpft}. In the case of BitFit and LP (\ie, the first step of LP-FT), bias and classifier layers fully utilize fine-tuning, while other weight layers (attention and MLP) rely on pre-trained models. We present an additional analysis in the Appendix~\ref{suppl:analysis_merge_ratio}.

\noindent\textbf{On $N$ fine-tuned models.} 
We further extend the previous derivation to $N$ fine-tuned models to move even closer to the weight center. 
Let us denote $\mathbf{w}_{\text{avr}}^{(N)}$ as the $N$-averaged weight, $\sum_{i=1}^N \mathbf{w}_i / N$, and $\mathbf{w}_H^{(N)}$ as the weight in the span$(\mathbf{w}_0, \mathbf{w}_1, \ldots, \mathbf{w}_N)$ closest to the distribution's center.
Then, we derive the position of $\mathbf{w}_H^{(N)}$ as:
\begin{equation}
    \begin{split}
        \mathbf{w}_H^{(N)} = & t \cdot \mathbf{w}_\text{avr}^{(N)} + (1-t) \cdot \mathbf{w}_0, \quad \quad \text{s.t.} \quad t = \frac{N\cos\theta}{1+(N-1)\cos\theta}.
    \end{split}
    \label{eq:interpolation_ratio}
\end{equation}
Detailed proof is in the Appendix~\ref{suppl:n_proof}. 
Similar to the case of two fine-tuned models, the interpolation ratio $t$ depends solely on the angle $\theta$ between the pre-trained model and the $N$ fine-tuned models.

\begin{wrapfigure}{r}{0.35\textwidth}
\vspace{-4.6em}
  \begin{center}
    \includegraphics[width=0.35\textwidth, height=2.5cm, keepaspectratio]{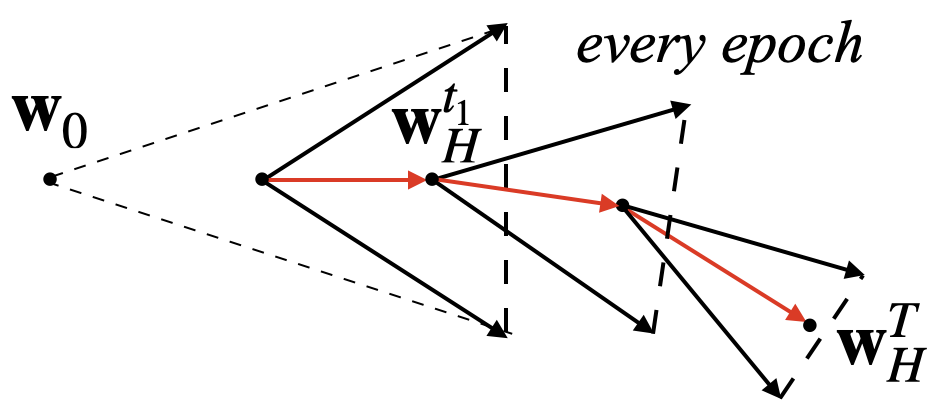}
  \end{center}
  \caption{\textbf{Periodic merging for \ours}. Weights are merged every epoch.}
  \label{fig:periodic_merging}
   \vspace{-3em}
\end{wrapfigure}
\noindent\textbf{Periodic merging.}
To move one step forward with our method, we propose \textit{periodic merging}, which is performed between the fine-tuned models and the pre-trained model during training.
As the geometric properties of weights are also applicable to weights during training (refer to Appendix~\ref{supp:during_training} 
for more details), we fully utilize this phenomenon here. 

We strategically merge weights at the end of every epoch, allowing for the fine-tuning process to be parallelized with merging.
The interpolation ratio for merging is determined by %
the angle $\theta$ between the pre-trained model and the fine-tuned models at the current epoch.
Fig.~\ref{fig:periodic_merging} visualizes this periodic merging process.
We argue that employing a periodic merging can approximate the center of weights more accurately.
In \S\ref{sec:exp_ablation}, we empirically show that the periodic merging yields superior performance and achieves a closer distance to the center.%

\section{Experiment}
\label{sec:exp}

We present the key experimental results in this section. 
We first provide our experimental setups in \S\ref{exp:setup}. Then, we present the main results in \S\ref{exp:main_result}, and ablation studies in \S\ref{sec:exp_ablation}.
More detailed experimental setups, analysis, and further results are in the Appendix~\ref{suppl:exp_setting}. 
We will release our codes and weights publicly.

\subsection{Experimental Setup}
\label{exp:setup}

\subsubsection{Models.}
We conduct the experiments on CLIP ViT-B/32, CLIP ViT-B/16, and CLIP ViT-L/14 models. 
We set the number of fine-tuning models for \ours as two.
We compare \ours against various fine-tuning techniques, including Model Soups~\cite{modelsoup}, LP-FT~\cite{lpft}, CAR-FT~\cite{car_ft}, FTP~\cite{tian2023fast}, and FLYP~\cite{flyp}. 
We use CLIP ViT-B/32 for ablation studies.

\vspace{-1.5em}
\subsubsection{Datasets.}
We fine-tune models on the ImageNet-1K~\cite{imagenet} training dataset. 
We report the ImageNet-1K~\cite{imagenet} top-1 accuracy for evaluating in-distribution (ID) performance.
For distribution shift scenarios, we consider five out-of-distribution (OOD) benchmarks including ImageNet-V2~\cite{imagenet_v2}, ImageNet-R~\cite{imagenet_r}, ImageNet-Sketch~\cite{imagenet_sketch}, ImageNet-A~\cite{imagenet_a}, and ObjectNet~\cite{objectnet}. 
Since previous methods, except Model Soups~\cite{modelsoup}, have not been evaluated on ObjectNet, we omit the ObjectNet results when comparing against them.

\vspace{-1.5em}
\subsubsection{Training setup.}
We initialize the classifier weights (\eg, 1000 classes for ImageNet) using the text encoder of CLIP and text prompts following Model Soup~\cite{modelsoup}. 
We use AdamW~\cite{adamw} with batch size 512 and weight decay 0.1 for all the experiments, including vanilla fine-tuning, Model Soup reproduction, and \ours\footnote{We reduce the batch size to 64 on CLIP ViT-L/14 due to memory limitation.}. 
We employ two training setups for comparisons: (1) training 10 epochs with minimal data augmentation, following Model Soup's zero-shot initialization setup, and (2) training 16 epochs with strong data augmentation, following Model Soup's LP initialization setup~\cite{modelsoup}, denoted with $^\star$. %
These setups enable a balanced comparison of \ours against zero-shot and LP-initialized Model Soups.
Detailed hyper-parameters are in the Appendix~\ref{suppl:exp_setting}.

\begin{table}[t]
\centering
\caption{
\textbf{Comparison against Model Soups~\cite{modelsoup} on CLIP ViT-B/32.}
We report the performance and relative fine-tuning costs on the CLIP ViT-B/32 scenario. $\alpha$ denotes the cost for LP initialization. 
\ours shows comparable performance with Model Soups with significantly reduced training costs. 
}
\label{tab:exp_vit_b32_soups}
\vspace{-1em}
\tabcolsep=0.75em
\resizebox{0.95\linewidth}{!}{
\begin{tabular}{@{}lccc@{}}
\toprule
Method                                     & ImageNet  & Avg. shifts & Cost \\ \midrule
{\color[HTML]{7A7A7A}{\textit{Comparing with Model Soups from zero-shot init.}}} & & & \\
CLIP zero-shot Initialization             & 63.34    & 48.51            & 0    \\
Vanilla FT                      & 78.35    & 47.03            & 1    \\
Uniform Model Soup (from zero-shot)                        & 79.76    & {\textbf{52.08}}            & 48   \\
Greedy Model Soup  (from zero-shot)                         & {\textbf{80.42}}    & 50.83            & 48   \\
\rowcolor{Gray} \ours                     & \underline{79.89}    & \underline{50.99}            & 2    \\
\midrule
{\color[HTML]{7A7A7A}{\textit{Comparing with Model Soups from LP init.}}} & & & \\
CLIP LP initialization                     &   75.57	       &    47.21              & $\alpha$    \\ 
Vanilla FT$^{\star}$                      & 79.72    & 46.37            & 1    \\
Uniform Model Soup (from LP init)                       & 79.97    & \textbf{{51.45}}            & 71+$\alpha$ \\
Greedy Model Soup  (from LP init)                     & \underline{81.03}    & \underline{50.75}            & 71+$\alpha$ \\
\rowcolor{Gray} \ourss                     & {\textbf{81.19}}    & {48.69}            & 2    \\ \bottomrule
\end{tabular}
}
\end{table}

\subsection{Main Results}
\label{exp:main_result}

\subsubsection{CLIP ViT-B/32.}
Table~\ref{tab:exp_vit_b32_soups} shows the results of \ours on the pre-trained CLIP ViT-B/32 model by comparing it with Model Soups. `Avg. shifts' denotes the average accuracy of the five OOD benchmark scores. 
Our \ours and \ourss show competitive performance with Model Soups. Furthermore, 
\ourss achieves state-of-the-art performance on ImageNet with \textbf{81.19}\% top-1 accuracy. 
As described in Fig.~\ref{fig:teaser}, \ours with WiSE-FT~\cite{wiseft} enjoys a superior ID-OOD performance curve compared to Model Soup and its WiSE-FT curves.
Note that Model Soups require dozens of fine-tuned models (\eg, zero-shot and LP-init Model Soups use 48 and 71 models, respectively), highlighting the effectiveness of \ours along with efficiency utilizing only two models. 
We provide further comparison results with WiSE-FT~\cite{wiseft} curves on LP-init Model Soups in the Appendix~\ref{suppl:additional_exp}.

\begin{table}[t]
\centering
\small
\caption{\textbf{\ours on CLIP ViT-B/16.} \ours shows competitive performance against previous fine-tuning methods on ImageNet and distribution shifts. 
}
\tabcolsep=.5em
\vspace{-1em}
\resizebox{0.8\linewidth}{!}{
\begin{tabular}{@{}lcccccc@{}}
\toprule
         &           & \multicolumn{5}{c}{Distribution shifts} \\ \cmidrule(l){3-7} 
\multirow{-2}{*}{Method}    & \multicolumn{1}{c}{ImageNet} & \multicolumn{1}{c}{Avg. shifts} & \multicolumn{1}{c}{IN-V2}   & \multicolumn{1}{c}{IN-R}    & \multicolumn{1}{c}{IN-A}    & \multicolumn{1}{c}{IN-Sketch} \\ \midrule
Zero-shot                       & 68.3      & 59.5         & 62.0     & \textbf{77.7}     & \underline{49.9}     & 48.3       \\
Vanilla FT                      & 82.8      & 57.7         & 72.9     & 66.4     & 43.7     & 48.0       \\
Vanilla FT$^\star$              & 83.7      & 57.4         & 73.5     & 67.6     & 40.0     & 48.6       \\
LP~\cite{lpft}                  & 79.7      & 48.1         & 71.5     & 52.4     & 27.8     & 40.5       \\
LP-FT~\cite{lpft}               & 81.7      & \underline{60.5}         & 71.6     & \underline{72.9}     & 49.1     & 48.4       \\
CAR-FT~\cite{car_ft}             & 83.2      &  59.4        & 73.0     & {71.3}     & 43.7     & {49.5}       \\
FTP~\cite{tian2023fast}         & \underline{84.2}    & {49.7}     & {74.6} & {47.2} & {26.5} & {50.2}   \\
FLYP~\cite{flyp}                 &  82.6       & \underline{60.5}       &  73.0     &  71.4       &  48.1     &   49.6   \\
\midrule
\rowcolor{Gray} \ours     & {84.1}    & \textbf{62.4}         & \underline{74.8}     & 71.8     & \textbf{51.2}     & \textbf{51.8}       \\
\rowcolor{Gray} \ourss   & \textbf{85.2}     & 60.1         & \textbf{75.3}     & 68.7     & 45.0     & \underline{51.3}      \\
\bottomrule
\label{exp:vit_b_16}
\end{tabular}
}
\vspace{-2em}
\end{table}

\vspace{-1em}

\subsubsection{CLIP ViT-B/16.}
Table~\ref{exp:vit_b_16} presents a comprehensive comparison of different fine-tuning methods applied to CLIP ViT-B/16.
Previous works \cite{lpft,tian2023trainable,tian2023fast} lack ObjectNet~\cite{objectnet} results; therefore, we omit ObjectNet and report the other four OOD benchmarks. 
Complete results with ObjectNet and ImageNet-Real~\cite{are_we_done} are in the Appendix~\ref{suppl:vit-b16_complete}.
The results show \ours exhibits exceptional performance on the ImageNet accuracy, \eg, \ourss achieves 85.2\% top-1 accuracy on ImageNet, which is a state-of-the-art level. 
\ours also shows robust performance across diverse distribution shift scenarios.

\begin{wraptable}{r}{0.38\textwidth}
\centering
\vspace{-3.3em}
\caption{\textbf{\ours on CLIP ViT-L/14.}}
\resizebox{0.85\linewidth}{!}{
\begin{tabular}{@{}lcc@{}}
\toprule
 & \multicolumn{1}{c}{IN} & \multicolumn{1}{c}{Avg. shifts} \\ \midrule
Zero-shot     & 75.0                         & 63.0                            \\
Vanilla FT        & 85.8                         & 66.8                            \\
Vanilla FT$^{\star}$    & 87.1                         & 68.0                            \\
TPGM~\cite{tian2023trainable}          & 87.0                         & 69.4                            \\
CAR-FT~\cite{car_ft}    & 87.1                         & 67.8      
\\
\midrule
\rowcolor{Gray} \ours   & {87.0}                         & {71.6}                            \\ 
\rowcolor{Gray} \ourss   & \textbf{87.7}                         & \textbf{73.5}                            \\ \bottomrule
\label{exp:vit_l_14}
\end{tabular}
}
\vspace{-2.5em}
\end{wraptable}

\vspace{-1em}
\subsubsection{CLIP ViT-L/14.}
Table~\ref{exp:vit_l_14} shows the results of \ours on the CLIP ViT-L/14 model.
The results show that \ours can push the limit of benchmark scores with large-size backbone architecture. 
We remark that \ourss achieves state-of-the-art performance with 87.7\% ImageNet top-1 accuracy, implying that \ours is still effective in a scale-up scenario.
The results consistently demonstrate the high efficacy and robustness of \ours across diverse scales of models and various benchmark scenarios, reaffirming its potential in practical applications.

\subsection{Ablation studies and analysis of \ours}
\label{sec:exp_ablation}
We conduct ablation studies on CLIP ViT-B/32. We train vanilla fine-tuned models and \ours for 16 and 8 epochs, respectively; thus, the training cost of \ours matches with a single fine-tuning process.

\begin{table}[t]
\centering
\small
\tabcolsep=0.5em
\begin{minipage}{0.47\textwidth}
    \centering
    \caption{\textbf{Impact of the number of fine-tuning models ($N$) on \ours.} IN denote ImageNet accuracy.}
    \vspace{-.5em}
    \resizebox{.95\linewidth}{!}{
    \begin{tabular}{@{}lccc@{}}
    \toprule
       & {IN} & {\small Avg. Shifts} & {$\|\mathbf{w} - \boldsymbol{\mu}\|$} \\ \midrule
    FT   & 79.7                        & 46.7                           & 13.13                    \\
    $N$=2        & 80.1                        & 48.8                           & 10.01                    \\
    $N$=3        & 80.2                        & 48.8                           & 9.05                     \\
    $N$=4        & \textbf{80.4}                        & \textbf{48.9}                           & 8.45                     \\ \bottomrule
    \label{table:abl_n_models}
    \end{tabular}
        }
\end{minipage}
\quad 
\begin{minipage}{0.47\textwidth}
    \vspace{-.7em}
    \centering
    \caption{\textbf{Impact of merging period on \ours.} IN denotes ImageNet accuracy.}
    \begin{tabular}{@{}lcc@{}}
    \toprule
    Period       & {IN} & \multicolumn{1}{l}{Avg. Shifts} \\ \midrule
    1000 iters & 79.8                        & 48.7                           \\
    5000 iters & 79.9                        & 48.5                           \\
    1 epoch   & \textbf{80.1}                        & \textbf{48.8}                           \\ \bottomrule
    \label{exp_abl_period}
    \end{tabular}
\end{minipage}
\vspace{-1em}
\end{table}

\vspace{-1em}
\subsubsection{Experiments on the number of fine-tuned models $N$.} 
Table~\ref{table:abl_n_models} shows the effect of the number of fine-tuned models.
The results show that \ours obtains enhanced performance and closer distance from the (pseudo-) center ({$\|\mathbf{w} - \boldsymbol{\mu}\|$}) as the number of merging models increases. 
Considering the trade-off between the performance and training cost induced by increased $N$, our setting ($N$=2) shows the best for \ours.

\vspace{-1em}
\subsubsection{Study on the merging period of \ours.}
Table~\ref{exp_abl_period} shows the results of various merging periods, including 1000 and 5000 iterations settings.
Note that $1$ epoch is $\sim$2500 iterations in our experiment. 
\ours shows consistent performance with various periods.

\vspace{-1em}
\subsubsection{The post-training merging strategy of \ours.}
We study an alternative of \ours that merges fine-tuned weights only once after each fine-tuning process is finished, similar to Model Soups~\cite{modelsoup}. 
We denote it as \ours (post-training). %
We utilize the individually fine-tuned weights as we conducted in \S\ref{sec:intuition}, using the same training settings with different random seeds for each model.
We report the performance and distance from the pseudo-center of \ours (post-training) in Table~\ref{tab:abl_post_merge}.
On the left side of the table, we provide the performance of its counterpart, a uniform averaging of $N$ models ($\mathbf{w}_\text{avg}^{N}$). The improvements from the uniform averaging to \ours (post-training) are denoted in the table's parentheses. 
The results show that \ours (post-training) archives improved distribution shift scores with closer distances toward the center than its counterpart.

\begin{table}[t]
\centering
\caption{\textbf{Post-training merging strategy of \ours.} 
We present ImageNet accuracy, distribution shifts, and distance from the center with the results of uniform averaging, a straightforward baseline. 
}
\tabcolsep=0.5em
\resizebox{\linewidth}{!}{
\begin{tabular}{@{}cccccccc@{}}
\toprule
 & \multicolumn{3}{c}{Uniform averaging ($\mathbf{w}_\text{avg}^{N}$)} &  & \multicolumn{3}{c}{\ours (post-training)}    \\ \cmidrule(l){2-4} \cmidrule(l){6-8} 
    & ImageNet     & Avg. Shifts     & $\|\mathbf{w} - \boldsymbol{\mu}\|$   &  & ImageNet       & Avg. Shifts       & $\|\mathbf{w} - \boldsymbol{\mu}\|$       \\ \cmidrule(l){2-4} \cmidrule(l){6-8} 
$N$=2 & 80.2         & 47.8           & 9.19     &  & {80.3\greenpscript{+0.1}} & \textbf{50.4\greenpscript{+2.6}}          & \textbf{7.62\greenpscript{-1.57}}  \\
$N$=3 & 80.4         & 48.2           & 7.44    &  & 80.4\greenpscript{+0.0}        & \textbf{50.2\greenpscript{+2.0}}          & \textbf{6.49\greenpscript{-0.95}}  \\
$N$=4 & 80.5         & 48.5           & 5.63    &  & 80.5\greenpscript{+0.0}        & \textbf{49.8\greenpscript{+1.4}}           & \textbf{5.16\greenpscript{-0.47}}  \\ \bottomrule
\end{tabular}
\label{tab:abl_post_merge}
}
\vspace{-1em}
\end{table}

\section{Related Work}
We discuss related works and highlight how our method differs and contributes to the existing works.

\vspace{-1em}
\subsubsection{Model Soups~\cite{modelsoup}} is a straightforward weight averaging method that merges weights from various fine-tuned models trained with different hyper-parameters. It demonstrates improved in-distribution (ID) and out-of-distribution (OOD) performance. While effective, model soups typically require a large number of fine-tuned models. Our method aims to achieve similar or superior performance improvements more efficiently, utilizing significantly fewer fine-tuning costs. 
We provide further discussion about Model Soups with our new interpretation in Appendix~\ref{suppl:discussion}.

\vspace{-1em}
\subsubsection{Robust Fine-tuning.}
When fine-tuning generalist models like CLIP~\cite{clip}, we often observe the fine-tuned models lose the generalization ability of the original ones, with decreased OOD performance. 
To address this issue, several robust fine-tuning approaches have been proposed. 
LP-FT~\cite{lpft} attempts to preserve pre-trained weights by initially training only a linear probing layer. WiSE-FT~\cite{wiseft} improves OOD performance through linear interpolation between fine-tuned and pre-trained weights. While our method shares similarities with WiSE-FT in using pre-trained weights, our method determines interpolation ratios layer-wise based on geometric properties.
Focusing on OOD performance, methods suggesting improved training objectives~\cite{car_ft,carot,lipsum-ft,flyp,tian2023trainable,tian2023fast} have been proposed. 
Our approach differs from these methods as we do not propose a new fine-tuning loss. Instead, we perform two fine-tunings and achieve robust performance by merging them. 

\subsubsection{Weight Center and Flat minima.}
Recent machine learning research has extensively explored the significance of finding flat minima for improved generalization~\cite{maddox2019simple, izmailov2018averaging, li2018visualizing, cha2021swad}. 
Keskar \etal~\cite{keskar2017on} and Hochreiter \& Schmidhuber~\cite{hochreiter1997flat} demonstrated that sharp optima by large batch SGD have steep, harmful directions, while broader optima enhance generalization. 
Stochastic Weight Averaging (SWA)~\cite{izmailov2018averaging} targets the center of flat minima, enhancing robustness against shifts in the loss landscape between training and test datasets. 
SWAG~\cite{maddox2019simple} builds on SWA by incorporating Bayesian model averaging with a Gaussian posterior to further boost performance.
SWAD~\cite{cha2021swad} found that the generalization gap between flat and sharp minima is more pronounced in OOD scenarios than in ID ones. 
Our method theoretically extends these approaches by efficiently identifying the center of flat minima with novel geometric properties, leading to significant improvements in both ID and OOD performances.

\subsubsection{Model Weight Merging.}
Recent research has explored merging models fine-tuned on various tasks. Methods such as Task Arithmetic~\cite{ilharco2023editing} and TIES~\cite{yadav2023tiesmerging} have been proposed. They are also based on the difference between fine-tuned and pre-trained weights (often referred to as the ``task vector''). However, our method distinguishes itself through geometric analysis for weight merging.
In the domain of Large Language Models (LLMs), merging techniques like WARM~\cite{rame2024warm} and WARP~\cite{ramé2024warp} have emerged. Our method has the potential for extension to these areas, offering new avenues for future research. 
\section{Conclusion}

Our study illuminated the fine-tuning process in machine learning, revealing that fine-tuned models' weights generally exhibit the properties of a Gaussian distribution. The proximity of these models to the center of weights was crucial for improved performance in target domains like ImageNet and under diverse distribution shifts. 
Utilizing a pre-trained model as a robust anchor point, we efficiently minimized the variance with fewer fine-tuned models, eliminating the need for additional training to find the optimal interpolation ratio.

Additionally, our findings suggested further knowledge and applicability to the models near flat minima and will offer new insights on model weight merging methods.
As the pretraining-finetuning paradigm gains more prominence, our insights will provide a foundation for better understanding and optimizing the fine-tuning process in both academia and industry.

\noindent\textbf{Limitation.} 
Due to resource limitations, we could not conduct larger-scale models such as ViT-G. Exploring this will be part of our future work.

\section{Acknowledgment}
We thank the researchers at NAVER AI Lab for their valuable comments. This work was supervised by Sangdoo Yun and Dongyoon Han. Dong-Hwan Jang is currently at Samsung Advanced Institute of Technology (SAIT).
\bibliographystyle{splncs04}
\bibliography{main}

\clearpage
\appendix

\renewcommand\thefigure{\Alph{figure}}    
\setcounter{figure}{0}  
\renewcommand\thetable{\Alph{table}}    
\setcounter{table}{0}  

{\large \noindent\textbf{Appendix}}
\vspace{2em}

\noindent In this Appendix, we provide in-depth analysis and additional insights to complement the main text of our study on \ours, our novel approach to fine-tuning and weight merging. 
The contents are summarized as follows: 
\begin{itemize}
    \item We examine the angle norm consistency of fine-tuned weights across various settings in \S\ref{suppl:consistency}, extending the observations discussed in \S\ref{subsec:geometric_prop_bw_weights}.
    \item We provide detailed proofs of geometric properties of fine-tuned weights in \S\ref{supp:proof_geometric}.
    \item We study the importance of reducing variance for performance in out-of-distribution scenarios in \S\ref{suppl:dist_shift}, showcasing the test error landscape across various datasets and elaborating on the explanations in \S\ref{subsec:weight_center}.%
    \item We provide detailed proofs in \S\ref{suppl:n_proof} for the optimal interpolation ratio in our method~\S\ref{sec:method}.
    \item We discuss prior studies through the lens of our findings in \S\ref{suppl:discussion}.
    \item We provide an additional analysis of the interpolation ratio in \S\ref{suppl:analysis_merge_ratio}.
    \item We present experimental settings of \S\ref{sec:exp} in \S\ref{suppl:exp_setting}. 
    \item We present additional experiments of \ours in \S\ref{suppl:additional_exp}
\end{itemize}
Each section aims to offer a comprehensive understanding of our method's underlying principles and its broad applicability in machine learning.

\section{Angle and Norm Consistency}
\label{suppl:consistency}
We argue that, as discussed in \S\ref{subsec:geometric_prop_bw_weights}, angles and norms of fine-tuned weights would remain consistent across fine-tuned models, independent of various factors. 
These factors include architecture type (ViTs~\cite{dosovitskiy2020vit}, ResNet~\cite{resnet}, ConvNeXt~\cite{convnext}), optimizers (SGD, AdamW~\cite{adamw}), augmentations (RRC~\cite{inception}, RandAug~\cite{randaug}), datasets (CIFAR~\cite{cifar}, ImageNet~\cite{imagenet}), or the initialization of the classifier (zero-shot, LP as in LP-FT~\cite{lpft}). We depict the layer-wise angle and norm of 5 fine-tuned weights for each category based on different random seeds.
We give detailed illustrations for each setting at the end of the Appendix to enhance readability 
(refer to Fig.~\ref{fig:angle_arch}--\ref{fig:angle_deit}). 
Across all these settings, the angle and norm of weights exhibit a surprising level of consistency.

\subsection{Analysis on layer-wise tendency}
The layer-wise angle and norm across various settings are shown in Fig.~\ref{fig:angle_arch}--\ref{fig:angle_init}. We visualize with every weight of attentions/convolutions (Attention/Conv), multi-layer perceptrons (MLP), normalizations (LayerNorm and BatchNorm), a classifier (Classifier), individual bias (Bias), and the remaining layers (\ie, the patchification layer, positional embedding, class embedding, and projection layer). We further display \textit{All} in each figure, which denotes the concatenation of the weights of entire layers.

The layer-wise analysis reveals an interesting trend: Bias and classifier layers demonstrate smaller angles than attention and MLP layers. 
In other words, bias and classifier layers exhibit lower randomness and more reliable updates than attention and MLP layers. It is important to note that as the angle decreases, the pre-trained model is less utilized for merging (refer to Eq.~\eqref{eq:two_inter_ratio}. 
This indicates that bias and classifier layers focus more on fine-tuned models and rely less on the pre-trained model, whereas attention and MLP layers depend less on the fine-tuned model (\ie, $t_{\text{bias}}, t_{\text{clf}} > t_{\text{attn}}, t_{\text{mlp}}$).
This observation extends the findings of previous works such as BitFit~\cite{zaken2021bitfit} and LP-FT~\cite{lpft}. 
In the case of BitFit and LP (\ie, the first step of LP-FT), bias and classifier layers fully utilize fine-tuning, while other layers (attention and MLP) rely on pre-trained models.

These traits could offer new insights into parameter-efficient transfer learning (PETL)~\cite{zaken2021bitfit, lian2022scaling, hu2021lora, he2021towards} and layer-wise fine-tuning~\cite{lee2022surgical, tian2023trainable, tian2023fast, lpft}. Maintaining weights with high randomness (higher angles) while updating on biases and classifier weights with lower randomness and fewer parameters would be an efficient fine-tuning strategy. PETL has been exploring this direction but has not yet provided solid reasons why certain layers are more effective than others. Our analysis suggests that one reason could be the lower randomness (or variance) of these layers, as indicated by the angle trend per layer.

\subsection{Maintaining consistency during training}
\label{supp:during_training}
We further argue that the consistency we observed is maintained while training progresses, as illustrated by multiple thin shells in Fig.~\ref{fig:gaussian}. %
To demonstrate that the angle and norm of fine-tuned models remain consistent during the entire training process, we plot their relationship across weights for every epoch in Fig.~\ref{fig:angle_during}. 
Please note that the angle is consistent across differently seeded models at the same timestamp (\ie, $\mathbf{w}_1|_{t=t_1}$ and $\mathbf{w}_2|_{t=t_1}$), not across models at different timestamps (\ie, $\mathbf{w}_1|_{t=t_1}$ and $\mathbf{w}_1|_{t=t_2}$).
The observed trend is as follows: as training progresses, the angle between weights steadily decreases. 
This analysis uses the CLIP ViT-B/32 model fine-tuned on ImageNet-1K with five random seeds.

\subsection{Filter-wise analysis of weights}
\label{suppl:filter_wise}
Li~\etal\cite{li2018visualizing} showed that when evaluating the robustness of a neural network by adding random noise to certain weights, performance analysis based on adding filter-wise noise (\ie, adding noise for each row in all weight matrices) aligns more closely with the generalization performance than adding layer-wise noise does.
Inspired by this observation, we investigate the possibility that the weight distribution may follow a filter-wise Gaussian distribution and adapt this concept to our method (see the performance analysis in~\S\ref{para:merging_unit}).
Fig.~\ref{fig:angle_filter} illustrates the angle distribution filter-wise.
The angle exhibits much larger standard deviations than the layer-wise distribution.
This could be attributed to the reduction in dimensionality.
As the number of dimensions decreases, it becomes challenging to approximate the norm as a constant value.

\subsection{Analysis on non-CLIP models}
To verify if this key observation also applies to non-CLIP models, we analyze the geometric patterns of fine-tuned weights trained using the DeiT~\cite{touvron2021deit} method (\ie, pre-trained on ImageNet-21K). Fig.~\ref{fig:angle_deit} displays the angle and norm of 10 DeiT-base models first pre-trained on ImageNet-21K~\cite{imagenet} and then fine-tuned on ImageNet-1K. We find that weights pre-trained with ImageNet-21K also exhibit \textit{consistent angle and norm}, indicating that our observation may be valid beyond CLIP fine-tuning scenarios as well.

\section{Detailed Proof for Geometric Properties of Fine-tuned Weights}
\label{supp:proof_geometric}

For all indices \( i, j \) within the set \( [1, N] \), where \( N \) denotes the sufficiently large number of fine-tuned weights, we derive one lemma and three propositions based on the foundational observation described in Eq.~\eqref{eq:obs}:

\noindent \textbf{Lemma:} \( \mathbf{w}_i \cdot \boldsymbol{\mu} = \boldsymbol{\mu} \cdot \boldsymbol{\mu} = l^2\cos\theta \).

\noindent\emph{Proof:}
\begin{align*}
\mathbf{w}_i \cdot \boldsymbol{\mu} &= \lim_{N\rightarrow \infty} \frac{1}{N} \mathbf{w}_i \cdot \sum_{k=1}^{N} \mathbf{w}_k = \lim_{N\rightarrow \infty} \frac{1}{N} (l^2 + (N-1)*l^2\cos\theta) \\
&= l^2\cos\theta.
\end{align*}

Similarly,
\begin{align*}
\boldsymbol{\mu} \cdot \boldsymbol{\mu} &= \lim_{N\rightarrow \infty} \frac{1}{N^2} \sum_{k=1}^{N} \mathbf{w}_k \cdot \sum_{l=1}^{N} \mathbf{w}_l = \lim_{N\rightarrow \infty} \frac{1}{N^2} (N*l^2 + N(N-1)*l^2\cos\theta) \\
&= l^2\cos\theta. \quad \qed
\end{align*}

\noindent \textbf{Proposition 1:} \( \| \mathbf{w}_i - \boldsymbol{\mu} \| = \text{constant} \).

\noindent\emph{Proof:}
\begin{align*}
\| \mathbf{w}_i - \boldsymbol{\mu} \|^2 &= (\mathbf{w}_i - \boldsymbol{\mu}) \cdot (\mathbf{w}_i - \boldsymbol{\mu}) \\
&= \mathbf{w}_i \cdot \mathbf{w}_i - 2\mathbf{w}_i \cdot \boldsymbol{\mu} + \boldsymbol{\mu} \cdot \boldsymbol{\mu} \\
&= l^2 - 2 l^2 \cos\theta + l^2\cos\theta \quad \text{(by Lemma)}\\
&= l^2 (1 - \cos \theta) \quad \text{(constant)} \quad \qed
\end{align*}

\noindent \textbf{Proposition 2:} \( (\mathbf{w}_0 - \boldsymbol{\mu}) \perp (\mathbf{w}_i - \boldsymbol{\mu}) \).

\noindent \emph{Proof:}
\begin{align*}
(\mathbf{w}_0 - \boldsymbol{\mu}) \cdot (\mathbf{w}_i - \boldsymbol{\mu}) &= -\boldsymbol{\mu} \cdot (\mathbf{w}_i - \boldsymbol{\mu}) \\
&= 0 \quad \text{(by Lemma)} \quad \qed
\end{align*}

\noindent \textbf{Proposition 3:} \( (\mathbf{w}_i - \boldsymbol{\mu}) \perp (\mathbf{w}_j - \boldsymbol{\mu}) \).

\noindent\emph{Proof:}
\begin{align*}
(\mathbf{w}_i - \boldsymbol{\mu}) \cdot (\mathbf{w}_j - \boldsymbol{\mu}) &= \mathbf{w}_i \cdot \mathbf{w}_j - \mathbf{w}_i \cdot \boldsymbol{\mu} - \mathbf{w}_j \cdot \boldsymbol{\mu} + \boldsymbol{\mu} \cdot \boldsymbol{\mu} \\
&= 0 \quad \text{(by Eq.~\eqref{eq:obs} \& Lemma)} \quad \qed
\end{align*}

\begin{figure}[t]
    \small
    \centering    
    \includegraphics[width=0.47\textwidth]{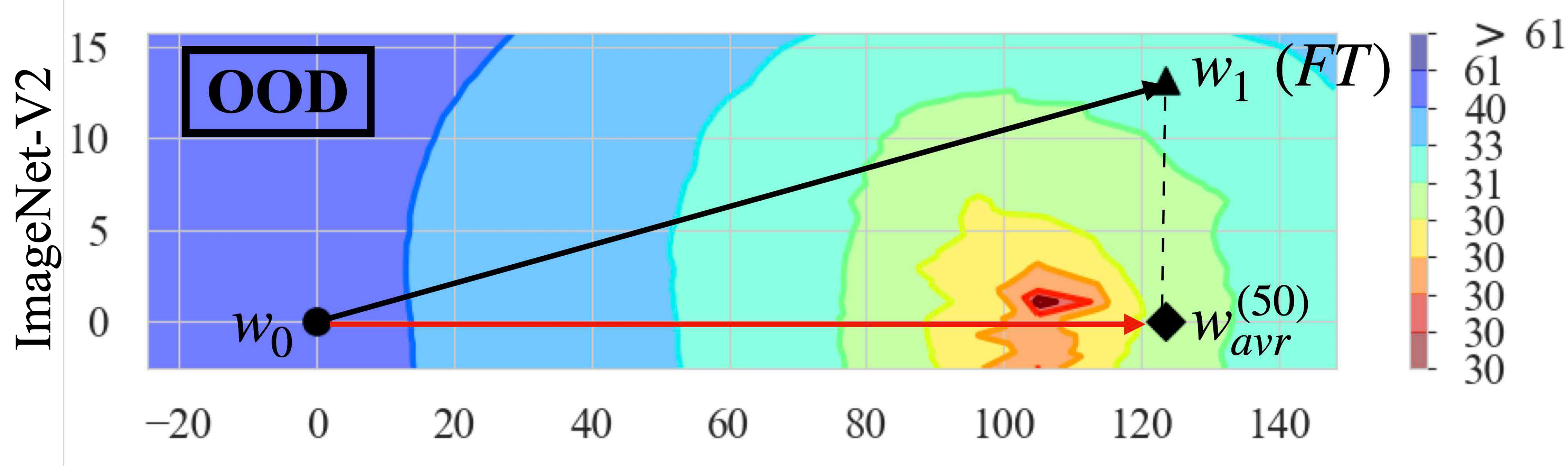}
    \includegraphics[width=0.47\textwidth]{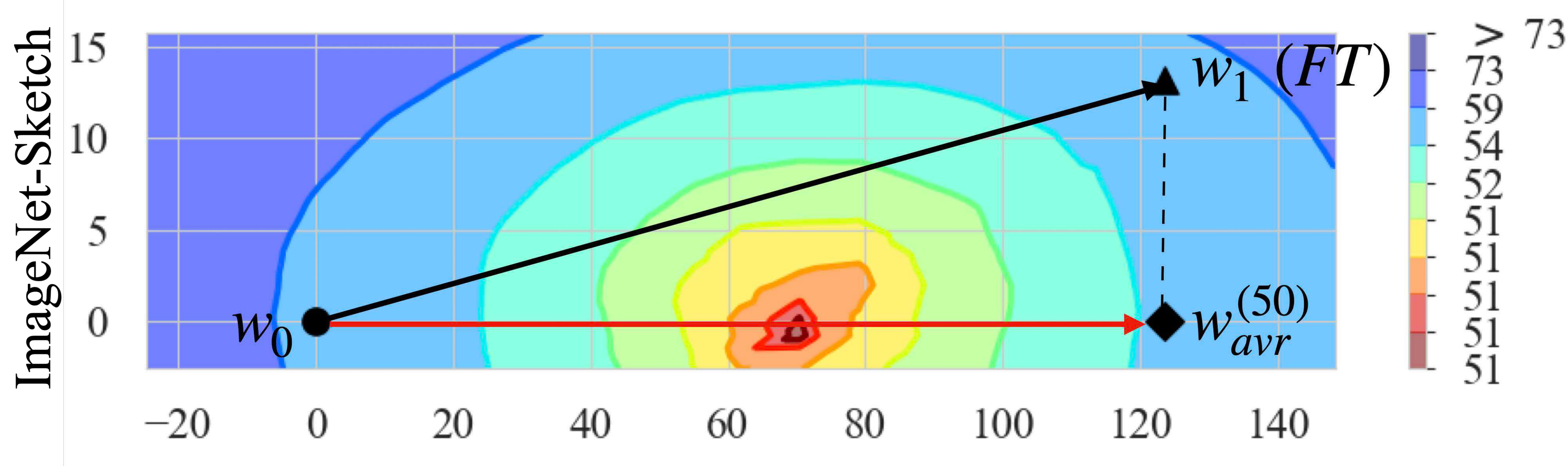}
    \includegraphics[width=0.47\textwidth]{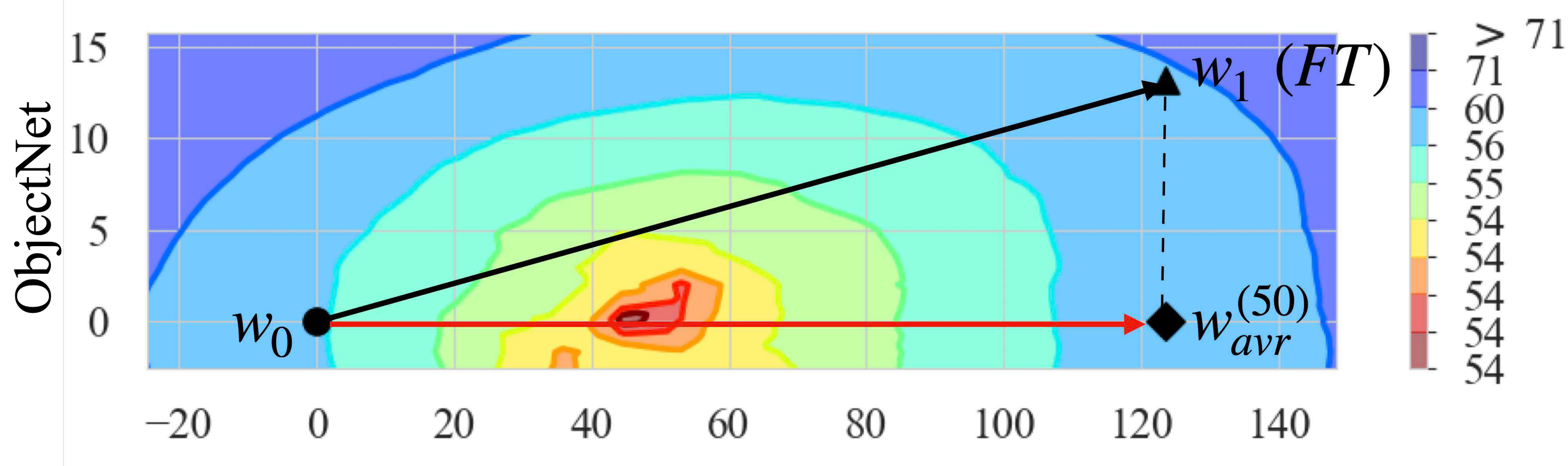}
    \includegraphics[width=0.47\textwidth]{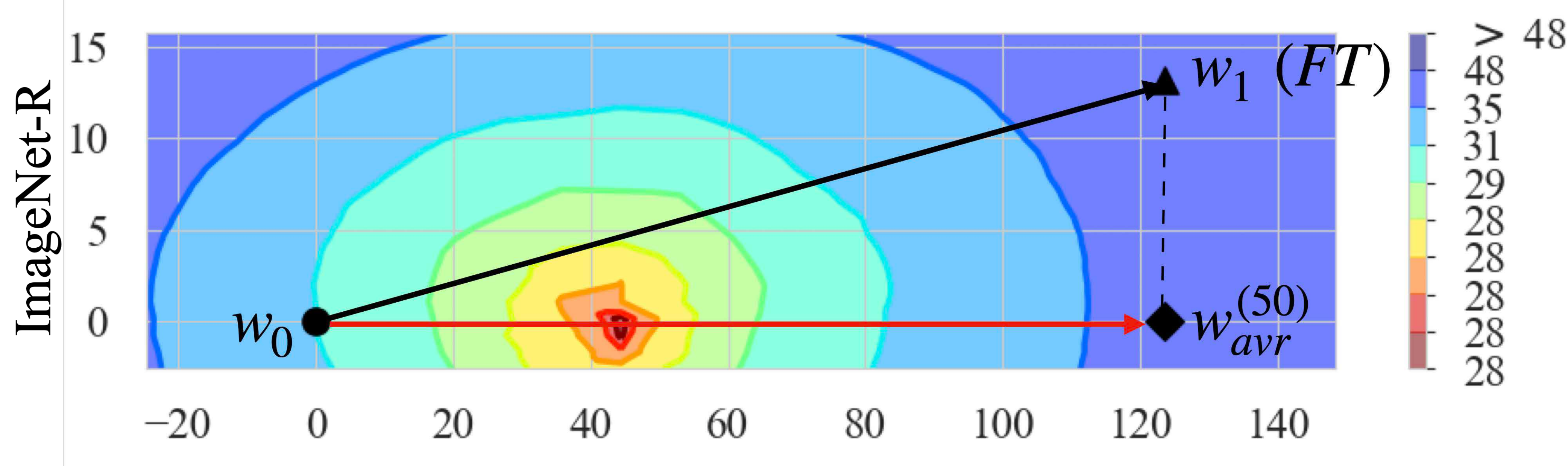}
    \includegraphics[width=0.47\textwidth]{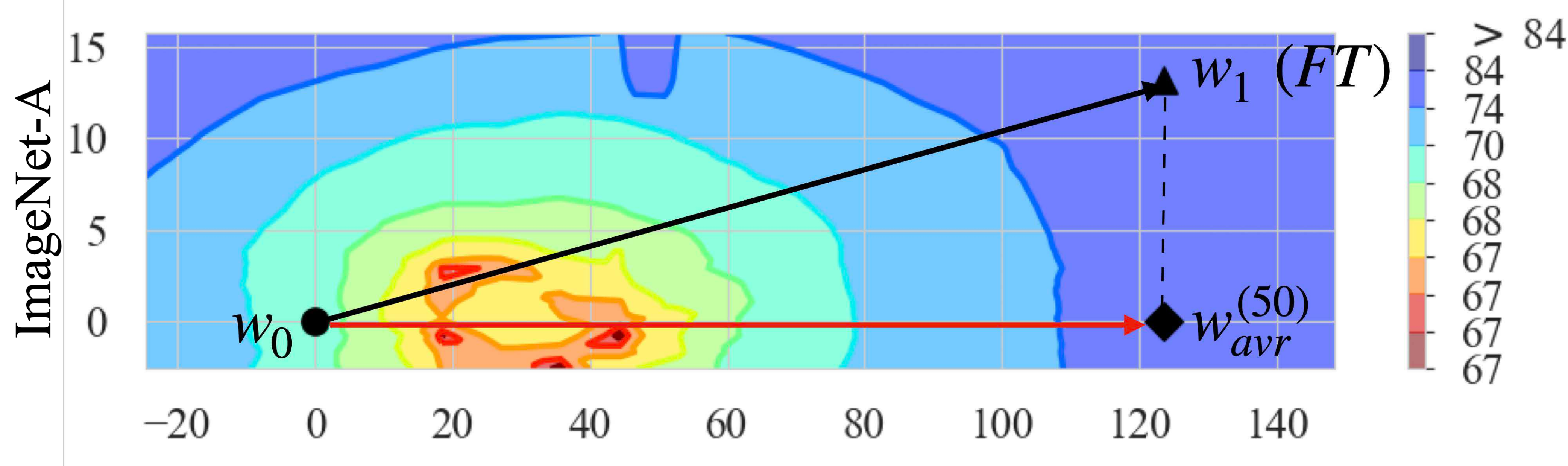}
    \includegraphics[width=0.47\textwidth]{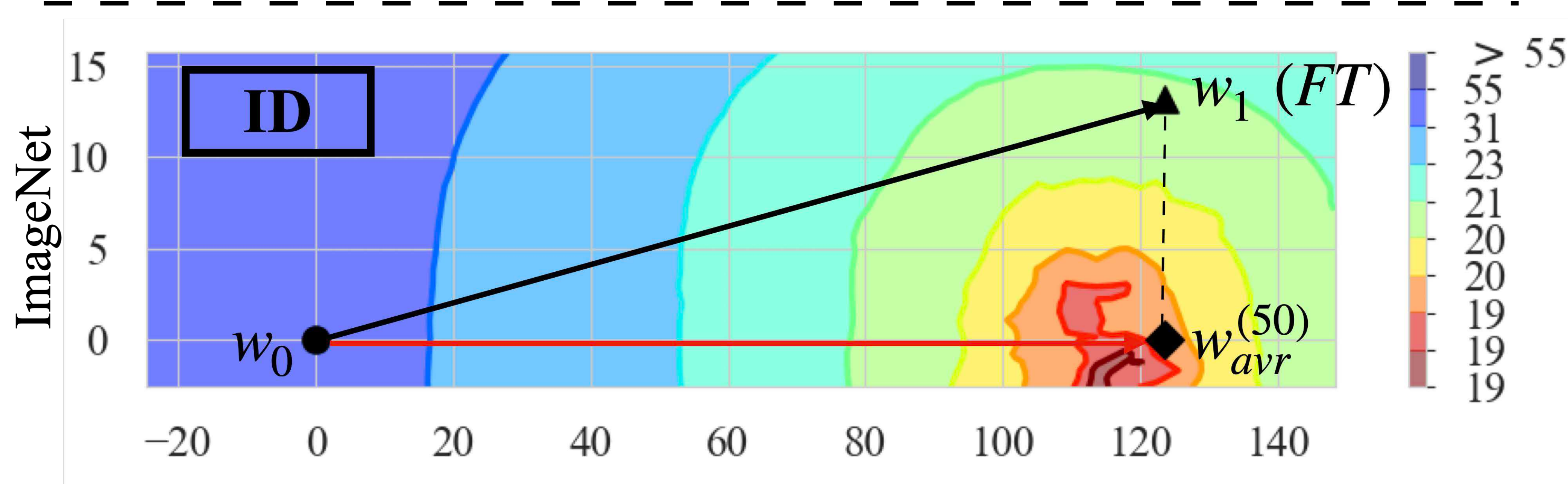}
    \vspace{-.5em}
    \caption{\textbf{Test error landscape on OOD datasets.} We depict the test error landscape on ImageNet-V2, -Sketch, ObjectNet, ImageNet-R, and -A (from left to right, from top to bottom, respectively) on the plane containing pre-trained model ($\mathbf{w}_0$), fine-tuned model ($\mathbf{w}_1$), and the pseudo-center of fine-tuned weights ($\mathbf{w}_{\text{avr}}^{(50)}$). The local optima for the OOD datasets always lie on the line segment $\overline{\mathbf{w}_0\mathbf{w}_{\text{avr}}^{(50)}}$.}
    \label{fig:landscape_ood}
    \vspace{-1.25em}
\end{figure}
\begin{figure}[t]
    \centering
    \small
    \includegraphics[width=0.65\textwidth]{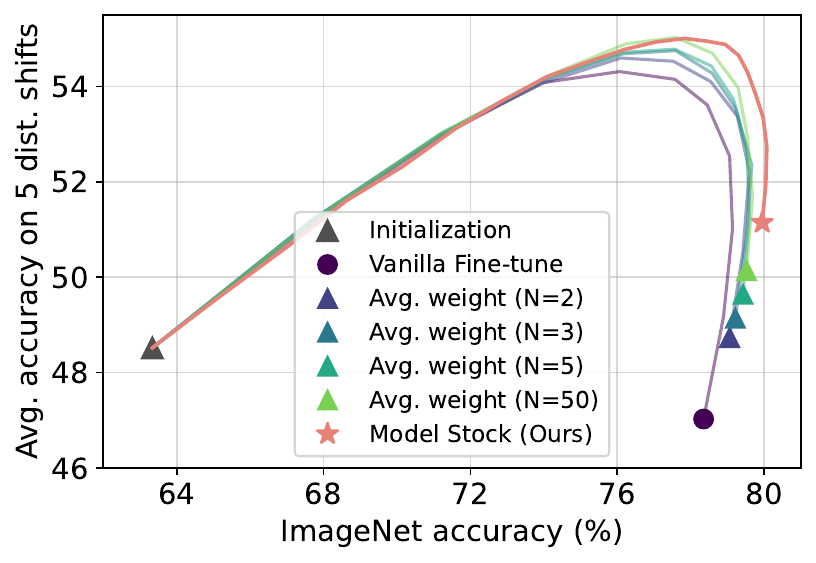} %
    \caption{\textbf{ID vs.\ OOD accuracy along WiSE-FT~\cite{wiseft} curves for averaged models.}
As the number of weights used for averaging increases, the corresponding WiSE-FT curves demonstrate improvements in the ID-OOD trade-off.}
    \label{fig:exp_id_ood_avr}
\end{figure}

\section{Importance of Reducing Weight Variance on Performance under Distribution Shifts}
\label{suppl:dist_shift}
In demonstrating the significance of variance reduction for robustness in out-of-distribution (OOD) scenarios, we analyze the test error landscape as in \S\ref{subsec:weight_center}. As shown in Fig.~\ref{fig:landscape_ood}, we examine the error landscape across various OOD datasets, including ImageNet-V2, ImageNet-Sketch, ObjectNet, ImageNet-R, and ImageNet-A (from top to bottom). This landscape is plotted on a plane defined by the weights of a pre-trained model ($\mathbf{w}_0$), a fine-tuned model ($\mathbf{w}_1$), and the center of the fine-tuned weights, which is approximated by averaging 50 fine-tuned weights ($\mathbf{w}_{\text{avr}}^{(50)}$). A notable pattern emerges where the local optima for these datasets consistently align with the line segment connecting $\mathbf{w}_0$ and $\mathbf{w}_{\text{avr}}^{(50)}$.

Though the exact location of local minima differs depending on the dataset type, it has a common point that the minima are aligned on the line between the weight center and pre-trained model rather than the line between the fine-tuned weight and pre-trained model.
Consequently, not only does the averaged weight exhibit higher performance on distribution shifts compared to the fine-tuned model, but the WiSE-FT~\cite{wiseft} curves corresponding to the averaged weights also demonstrate better ID/OOD trade-off than the WiSE-FT curve of the fine-tuned model, as illustrated in Fig.~\ref{fig:exp_id_ood_avr}.
This indicates the importance of getting closer to the weight center, even for OOD datasets.

Another interesting point is that depending on the traits of datasets, the position of local minima differs.
ImageNet-V2 has a similar dataset distribution to ImageNet since it shares the same data collection and categorization policy, and its local optima lies close to that of ImageNet.
On the other hand, on the datasets with harsh variations (\eg, ImageNet-A), the local minima are positioned much closer to the pre-trained model than the original ImageNet or ImageNet-V2.
This loss landscape gives an intuitive insight into the similarity between OOD datasets and ImageNet.

In conclusion, there is no universal interpolation ratio optimal for every distribution shift. 
However, all the local minima lie on the line between the weight center and the pre-trained model. 
This implies the importance of proximity to the weight center in achieving a better WiSE-FT line.

\section{Detailed Proof of \ours}
\label{suppl:n_proof}

Here, we present detailed proof of \ours introduced in \S{\ref{sec:method}}. 
We first show the case with two fine-tuned models and extend our proof toward $N$ fune-tuned models. 

\paragraph{On two fine-tuned models.}
We will prove step-by-step how the optimal interpolation ratio $t$ in Eq.~\eqref{eq:two_inter_ratio} in the main paper is derived.
Using the same notation as in \S\ref{sec:method}, we denote the magnitude and the angle between the fine-tuned weights as $l$ and $\theta$, respectively.
Starting from the fact that $\triangle \boldsymbol{\mu} \mathbf{w}_1 \mathbf{w}_2$ is a right isosceles triangle, we can derive the following relations from Fig.~\ref{fig:two_method_proof}:

\begin{figure}[t]
    \centering
    \includegraphics[width=0.55\textwidth]{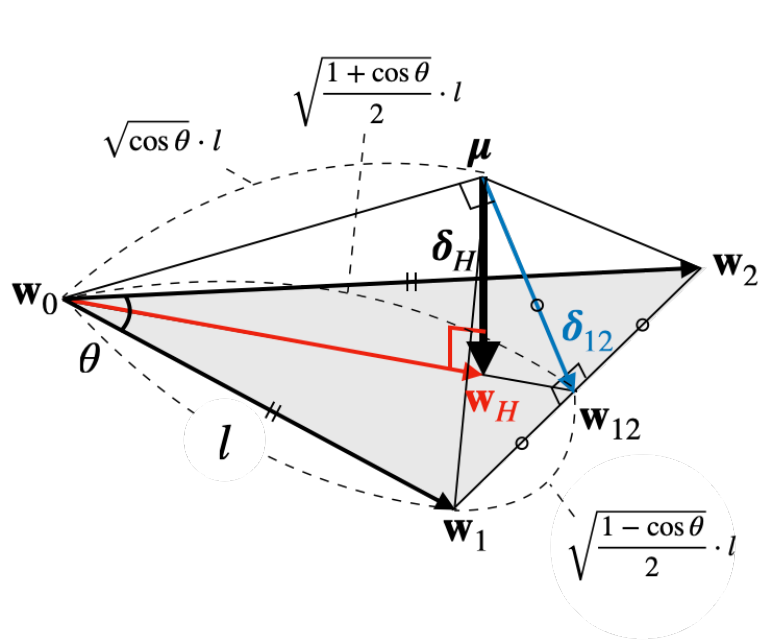}
    \vspace{-1em}
    \caption{\textbf{Model Stock with two fine-tuned models.} We reference the illustration in Fig.~\ref{fig:method_2_and_n} to more understandably substantiate merging two fine-tuned models.}
    \label{fig:two_method_proof}
    \vspace{-1.5em}
\end{figure}

\begin{align}
    \overline{\mathbf{w}_{12}\mathbf{w}_1} &= \overline{\mathbf{w}_{12}\mathbf{w}_2} = \overline{\mathbf{w}_{12}\boldsymbol{\mu}} \notag \\
    &= \sqrt{\frac{1-\cos\theta}{2}}\cdot l \quad \text{(from $\triangle \mathbf{w}_0 \mathbf{w}_1 \mathbf{w}_2$ and $\triangle \boldsymbol{\mu} \mathbf{w}_1 \mathbf{w}_2$)} \label{eq:w12w1}
\end{align}
\vspace{-1em}
\begin{align}
    \Rightarrow \overline{\mathbf{w}_{12}\mathbf{w}_0} &= \sqrt{\overline{\textbf{w}_1\textbf{w}_0}^2 - \overline{\mathbf{w}_{12}\mathbf{w}_1}^2} \notag \\
    &= \sqrt{1^2-\frac{1-\cos\theta}{2}}\cdot l \notag \\
    &= \sqrt{\frac{1+\cos\theta}{2}}\cdot l \quad \text{(from $\triangle \mathbf{w}_0 \mathbf{w}_1 \mathbf{w}_{12}$ and Eq.~\eqref{eq:w12w1})} \label{eq:w12w0}
\end{align}
\vspace{-1em}
\begin{align}
    \Rightarrow \overline{\mathbf{w}_0\boldsymbol{\mu}} &= \sqrt{\overline{\mathbf{w}_{12}\mathbf{w}_0}^2-\overline{\mathbf{w}_{12}\boldsymbol{\mu}}^2} \notag \\
    &= \sqrt{\frac{1+\cos\theta}{2} - \frac{1-\cos\theta}{2}}\cdot l \notag \\
    &= \sqrt{\cos\theta}\cdot l \quad \text{(from $\triangle \mathbf{w}_0 \boldsymbol{\mu} \mathbf{w}_{12}$, Eq.~\eqref{eq:w12w1} and Eq.~\eqref{eq:w12w0})} \label{eq:w0mu}
\end{align}
\vspace{-1em}
\begin{align}
    \Rightarrow t &:= \frac{\overline{\mathbf{w}_H\mathbf{w}_0}}{\overline{\mathbf{w}_{12}\mathbf{w}_0}} \notag = \frac{\overline{\mathbf{w}_H\mathbf{w}_0}}{\overline{\mathbf{w}_0 \boldsymbol{\mu}}} \cdot \frac{\overline{\mathbf{w}_0 \boldsymbol{\mu}}}{\overline{\mathbf{w}_{12}\mathbf{w}_0}} \notag \\
    &= \left(\frac{\overline{\mathbf{w}_0 \boldsymbol{\mu}}}{\overline{\mathbf{w}_{12}\mathbf{w}_0}}\right)^2 \quad \text{(from $\triangle \mathbf{w}_0 \boldsymbol{\mu} \mathbf{w}_{12}$ $\sim$ $\triangle \mathbf{w}_0 \mathbf{w}_H \boldsymbol{\mu}$)} \notag \\
    &= \frac{2\cos\theta}{1+\cos\theta}  \quad \text{(from Eq.~\eqref{eq:w12w0} and Eq.~\eqref{eq:w0mu})} \label{eq:two_t} 
\end{align}
\hfill $\qed$

Interestingly, $\mathbf{w}_H$ is located at an orthocenter of the triangle $\triangle \mathbf{w}_0 \mathbf{w}_1 \mathbf{w}_2$ with the given optimal ratio $t$.

\begin{figure}[t]
    \centering
    \begin{subfigure}[b]{0.5\textwidth}
        \includegraphics[width=\textwidth]{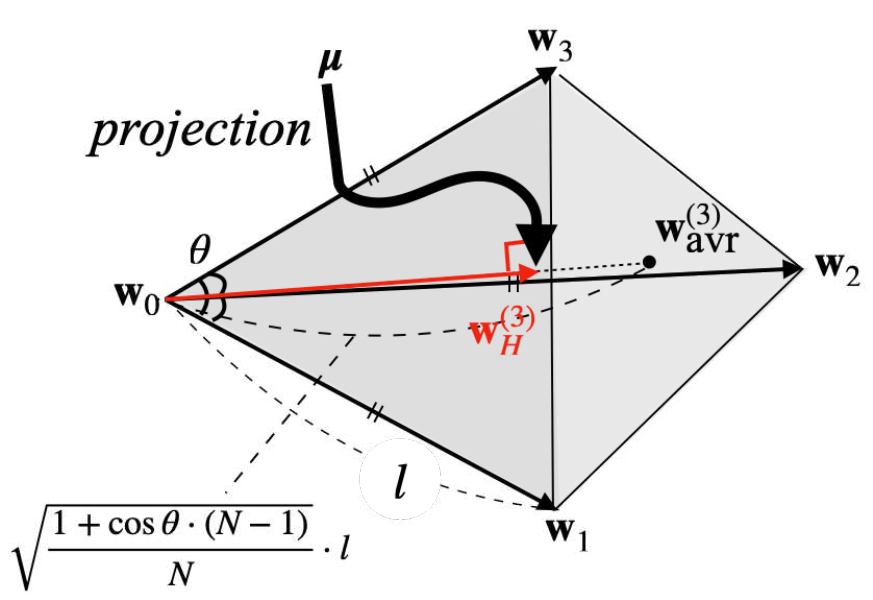}
        \caption{Model Stock with $N$ fine-tuned models.}
        \label{subfig:n_method_proof}
    \end{subfigure}
    ~ %
    \begin{subfigure}[b]{0.45\textwidth}
        \includegraphics[width=\textwidth]{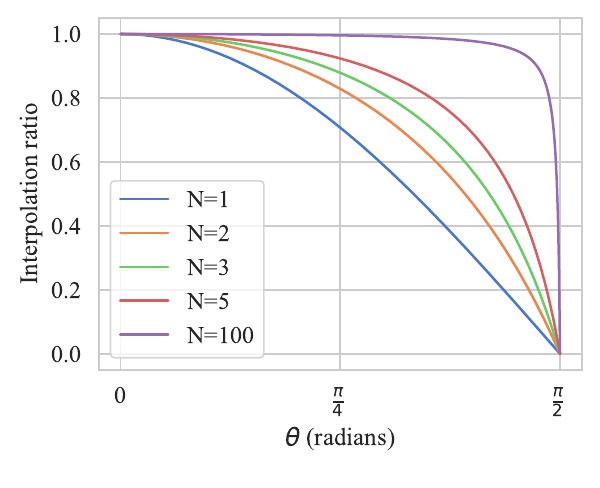}
        \caption{Variation of the interpolation ratio $t$.}
        \label{subfig:t_func}
    \end{subfigure}
    \caption{\textbf{Model Stock with $N$ fine-tuned models and Interpolation Ratio Variation.} (a) We visualize a special case of $N=3$ (tetrahedron) for better understanding. (b) The trend towards $t=1$ with increasing $N$ illustrates that $\mathbf{w}_H^{(N)}$ on the $N$-dimensional simplex approaches $\mathbf{w}_{\text{avr}}^{(N)}$, reflecting a growing dependence on the number of fine-tuned models.}
    \label{fig:merged_figures}
\end{figure}

\paragraph{On $N$ fine-tuned models.}
Similarly, we can derive a more generalized interpolation ratio for $N \geq 2$. 
Our goal is to find the weight $\mathbf{w}_\text{avr}^{(N)}$ that is on the hyper-plane spanned by $\mathbf{w}_0, \mathbf{w}_1, \ldots, \mathbf{w}_N$ and closest to the weight center $\boldsymbol{\mu}$, as described in Fig.~\ref{subfig:n_method_proof}.
Again, for simplicity, we treat $\mathbf{w}_0$ as the origin $\mathbf{O}$. 

Based on the observation, we presume that the following two conditions hold:
\begin{equation}
    \begin{cases}
        \mathbf{w}_H^{(N)} =  t \cdot \mathbf{w}_\text{avr}^{(N)} \\
        (\mathbf{w}_\text{avr}^{(N)}-\mathbf{w}_H^{(N)}) \cdot (\boldsymbol{\mu} - \mathbf{w}_H^{(N)}) = 0. \label{eq:n_condition}
    \end{cases}    
\end{equation}
The first condition comes from the symmetry of an $N$-simplex structure, and the second condition holds since the orthogonal projection is the minimal distance from $\boldsymbol{\mu}$.
Then, we can derive $t$ as follows:

By substituting the first condition into the second condition from Eq.~\eqref{eq:n_condition},
\begin{align}
    & (\mathbf{w}_\text{avr}^{(N)}-\mathbf{w}_H^{(N)}) \cdot (\boldsymbol{\mu} - \mathbf{w}_H^{(N)}) = 0 \notag \\
    \Rightarrow \; &  \mathbf{w}_\text{avr}^{(N)} \cdot \boldsymbol{\mu} - t \cdot \|\mathbf{w}_\text{avr}^{(N)}\|^2 = 0 \notag \\ %
    \Rightarrow \; & t = \frac{\boldsymbol{\mu} \cdot \mathbf{w}_{\text{avr}}^{(N)}}{\|\mathbf{w}_{\text{avr}}^{(N)}\|^2}.\label{eq:t_eq} %
\end{align}

Note that the norm of the $N$-averaged fine-tuned weights can be derived as follows:
\begin{align}
    \|\mathbf{w}_{\text{avr}}^{(N)}\|^2 &= \frac{1}{N^2} (\mathbf{w}_1 + \ldots + \mathbf{w}_N) \cdot (\mathbf{w}_1 + \ldots + \mathbf{w}_N) \notag \\
    &= \frac{1}{N^2} (l^2 + l^2 \cos\theta \cdot (N-1)) \cdot N \notag \\
    &= \frac{l^2}{N} (1 + \cos\theta \cdot (N-1)), \label{eq:w_avr_norm}
\end{align}
while the term $\boldsymbol{\mu} \cdot \mathbf{w}_{\text{avr}}^{(N)}$ can be simplified as 
\begin{equation}
    \boldsymbol{\mu} \cdot \mathbf{w}_{\text{avr}}^{(N)} = \frac{1}{N}\sum_{i=1}^{N}(\boldsymbol{\mu} \cdot \mathbf{w}_i) = l^2 \cos\theta \quad \text{(from Lemma)}. \label{eq:mu_wavr}
\end{equation}

By substituting Eq.~\eqref{eq:w_avr_norm} and Eq.~\eqref{eq:mu_wavr} into Eq.~\eqref{eq:t_eq}, we can finally derive the optimal interpolation ratio $t$ as follows:
\begin{equation}
    t = \frac{N\cos\theta}{1+(N-1)\cos\theta} \quad \qed
\end{equation}

Fig.~\ref{subfig:t_func} displays how the optimal interpolation ratio $t$ varies as a function of $\theta$ with different numbers of fine-tuned models.
As $N$ increases, $t$ trends towards 1, indicating that $\mathbf{w}_H^{(N)}$ on the $N$-dimensional simplex gets closer to $\mathbf{w}_{\text{avr}}^{(N)}$.
This shows increasing dependence on fine-tuned models as their number grows.

\section{Discussion --- Rethinking Pivotal Prior Studies}
\label{suppl:discussion}

\begin{figure}[t!]
    \centering
    \includegraphics[width=0.8\textwidth]{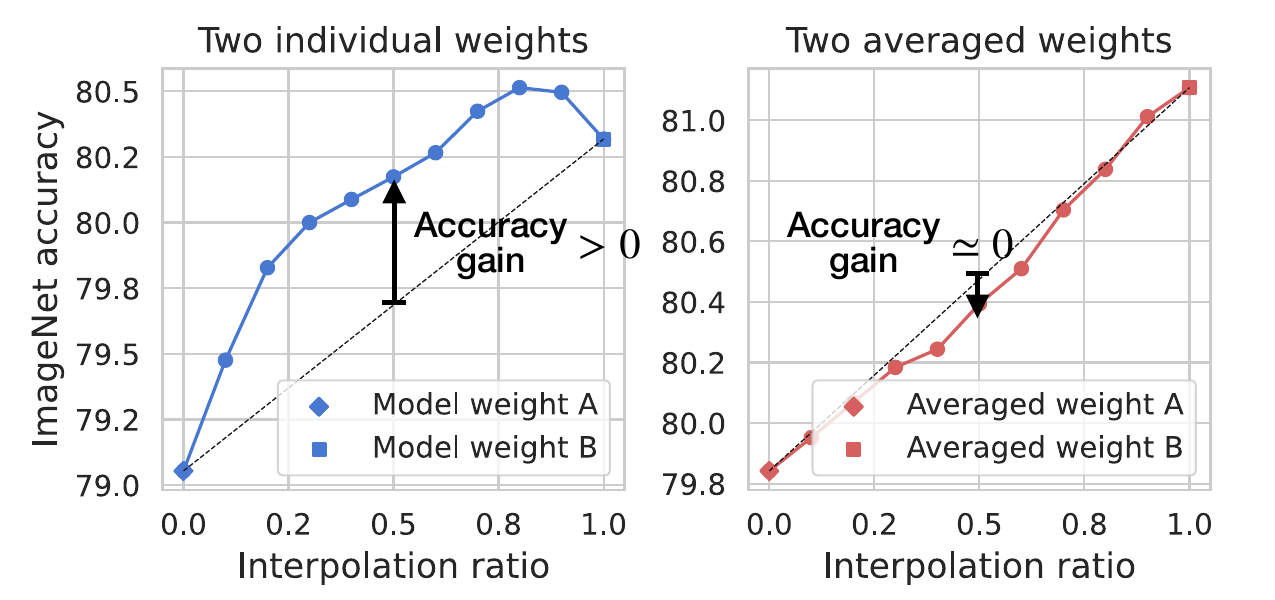} %
    \caption{%
    \textbf{Ensembling impact disappears when interpolating between two averaged weights.}     
    We plot the ImageNet performance of interpolated weights between two selected fine-tuned models in Model Soup~\cite{modelsoup} (left) and between their corresponding weight centers (right). 
    }
    \label{fig:model_soups}
\end{figure}

In this section, we extend our findings to reinterpret the underlying mechanics in prior studies,
WiSE-FT~\cite{wiseft} and Model Soups~\cite{modelsoup}, through a consistent rationale to illuminate their effectiveness.

\vspace{-1em}
\subsubsection{WiSE-FT}\cite{wiseft} is a state-of-the-art robust fine-tuning method for CLIP-based models. 
It demonstrates that linearly combining weights of the pre-trained and fine-tuned models achieves a significant accuracy gain on distribution shifts. %
We argue that the WiSE-FT model's superiority over a fine-tuned model can be interpreted by its weights being closer to the center of the corresponding weight distribution. 
Fig.~3 %
already showed fine-tuned models typically lie on the periphery of flat minima.
Given that the angle $\angle\textbf{w}_0\textbf{w}_{\text{avr}}^{(50)}\textbf{w}_1$ is nearly a right angle, along the line $\overline{\textbf{w}_0\textbf{w}_1}$, multiple weight points are closer to the center than a single fine-tuned model, thereby enhancing performance.
Note that $\textbf{w}_H$ is the closest to the center among the line $\overline{\textbf{w}_0\textbf{w}_1}$. %
More discussions on performance boosts observed in distribution shifts are provided in the Appendix~\ref{suppl:dist_shift}.%

\subsubsection{Model Soup}\cite{modelsoup} merges various fine-tuned models' weights trained from varied hyper-parameters. It has been credited with delivering enhanced performance across ImageNet and distribution shifts.
Here, we interpret the performance improvements of Model Soup as the result of the proximity to the center of weight distribution. 
Consider two weight vectors, $\mathbf{w}_A$ and $\mathbf{w}_B$, fine-tuned with different hyper-parameters and following Gaussian distribution $\mathcal{N}(\boldsymbol{\mu_A}, \Sigma_A)$ and $\mathcal{N}(\boldsymbol{\mu_B}, \Sigma_B)$ respectively.
Then, the interpolated weight vector $\mathbf{w}_{AB} = t \cdot \mathbf{w}_A + (1 - t) \cdot \mathbf{w}_B$ also follows a Gaussian distribution $\mathcal{N}(\boldsymbol{\mu_{AB}}, \Sigma_{AB})$.
The expected squared distance from the interpolated weight vector to its mean $\boldsymbol{\mu_{AB}}$ is minimized to $\frac{\text{trace}(\Sigma_A)\text{trace}(\Sigma_B)}{\text{trace}(\Sigma_A)+\text{trace}(\Sigma_B)}$ when $t$ is chosen to $\frac{\text{trace}(\Sigma_B)}{\text{trace}(\Sigma_A)+\text{trace}(\Sigma_B)}$, indicating the reduction of variance through weight interpolation (\ie, the distance between $\mathbf{w}_{AB}$ and $\boldsymbol{\mu_{AB}}$ might be closer than each weight's distance). 
For example, if $\text{trace}(\Sigma_B)$ is equal to $\text{trace}(\Sigma_A)$,  this minimum squared distance is exactly half of the sum of the individual traces when $t=0.5$.
This insight suggests that the performance gains realized by Model Soup could be due to reduced variance resulting from merging numerous weights.

We set up a toy experiment to evaluate the effect of variance reduction in the Model Soup scenario by comparing the interpolation of fine-tuned weights with the interpolation of their corresponding weight centers, when $N=2$. In the former case, variance reduction exists along with the effect of merging diverse hyper-parameters, while in the latter case, performance gain would only come from hyper-parameter diversity. 
If the diversity of hyper-parameters is a major factor, the performance gain from interpolation of central weights should remain the same.
To test this, we assessed the ImageNet performance of interpolated weights between pairs of fine-tuned models within Greedy Model Soup\footnote{We opt for Greedy Model Soup to show that even the interpolation of models from the best merging combination does not benefit from the impact of weight diversity.}~\cite{modelsoup} and compared it to interpolations between their central weights, calculated as the average of 20 differently seeded models. 
Fig.~\ref{fig:model_soups} shows that, unlike interpolations between individual models, using the centers does not significantly improve performance. 
This suggests that proximity to the center of the weight distribution may play a more critical role than hyper-parameter diversity in weight ensemble methods in this case.

It is also worth noting that $\boldsymbol{\mu_{AB}}$ always surpasses the performance of $\mathbf{w_{AB}}$ for the same interpolation ratio $t$, indicating that the importance of proximity to the center remains consistent for interpolated weights.
With extensive future research, this understanding could provide valuable insights for developing more generalizable and effective weight-merging techniques.

\begin{figure}[t]
    \centering
    \begin{subfigure}[b]{0.48\textwidth}
        \includegraphics[width=\textwidth]{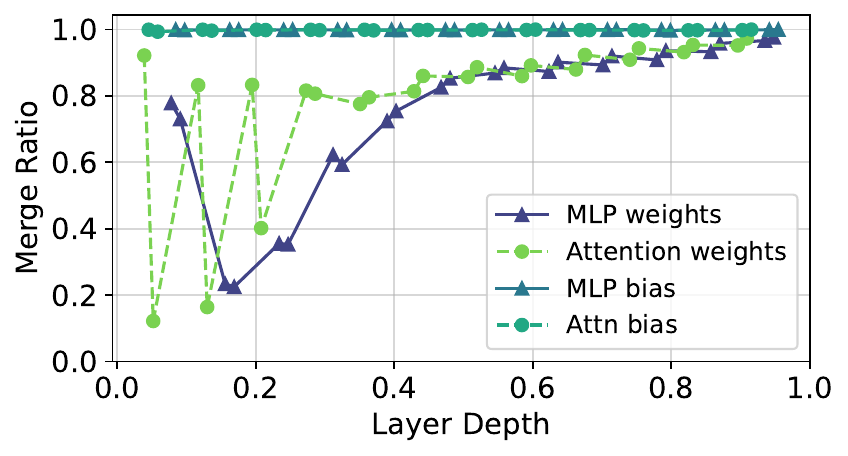}
        \caption{\textbf{Layer depth vs. interpolation ratio $t$.} } 
        \label{subfig:appendix_merge_ratio_depth}
    \end{subfigure}
    ~ %
    \begin{subfigure}[b]{0.48\textwidth}
        \includegraphics[width=\textwidth]{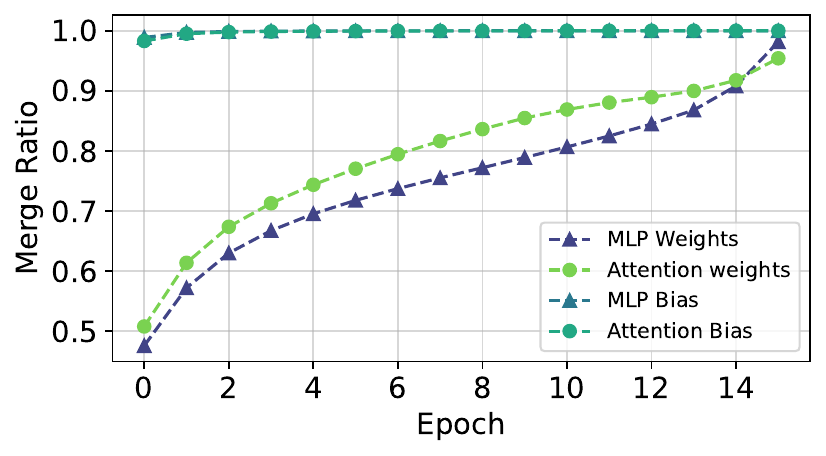}
        \caption{\textbf{Epoch vs. interpolation ratio $t$.} } 
        \label{subfig:appendix_merge_ratio_epoch}
    \end{subfigure}               
    \caption{\textbf{Trend of interpolation ratio $t$ during \ours training.}
    }
    \label{fig:supp:analysis_merge_ratio}
\end{figure}

\section{Analysis of the interpolation ratio $t$}
\label{suppl:analysis_merge_ratio}
We analyze the interpolation ratio $t=\frac{2\cos\theta}{1+\cos\theta}$ in a layer-wise manner.
During a \ours experiment on CLIP ViT-B/32 with 16 epoch training, we log the layer-wise merge ratios at every merging period. 
Figure~\ref{fig:supp:analysis_merge_ratio} visualizes the averaged interpolation ratio during \ours training. 
We plot two trends of the interpolation ratio for the layer depth and training step. 
Our overall observation indicates the bias layers have high merge ratios $t$ ($\simeq1$) with small angles $\theta$ ($\simeq0$), implying that the bias layers do not need to enjoy the pre-trained model, similar to our discussion in \S\ref{sec:intuition} and \S\ref{sec:method}. 
Focusing on the weight layers, Figure~\ref{subfig:appendix_merge_ratio_depth} shows a U-shape tendency as the layer depth increases, implying the weights of intermediate layers can be more diverse (\ie, larger angle $\theta$) than those of early and later layers.
Our intuition here is that since the early and later layers are directly connected to input data and output labels, respectively, they may not demand the advantage of the pre-trained weight. 
Figure~\ref{subfig:appendix_merge_ratio_epoch} presents that the models at the early training stage are more diverse and they enjoy the pre-trained weights more than those of the later training stage. As the model approaches convergence, the diversity of fine-tuning models decreases (\ie, smaller angle $\theta$).

\section{Experimental setup}
\label{suppl:exp_setting}

Here, we present detailed setups for the experiments in \S{\ref{sec:exp}}.
We utilize AdamW optimizer~\cite{adamw} with a weight decay of 0.1. 
We employ two training setups for \ours. 
The first is training \ours with a learning rate of $3\times10^{-5}$ in 10 epochs with minimal data augmentation.
The minimal data augmentation utilizes random resize crop augmentation with a minimum crop ratio of $0.9$, mixup~\cite{zhang2017mixup} augmentation with $\beta$=0.5, following Model Soup's ``standard grid search'' setting.  
The other is training \ours with a learning rate of $2\times10^{-5}$ in 16 epochs with strong data augmentation. 
The strong data augmentation utilizes random resize crop augmentation with a minimum crop ratio of $0.08$ and random augmentation~\cite{randaug} ($N=2$, $M=10$) following Model Soup's ``random search'' setting. 
When experimenting with the ViT-B/16 and ViT-L/14 models, we adjusted the learning rate and batch size to accommodate the GPU memory constraints.

\begin{figure}[t!]
    \centering
    \includegraphics[width=0.6\linewidth]{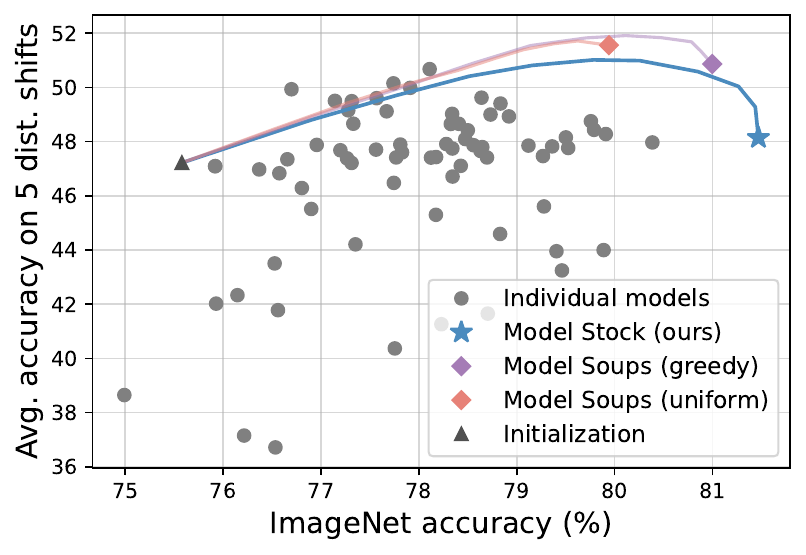}
    \caption{
    \textbf{Results on LP initialization}. 
    We plot in-distribution ImageNet accuracy (x-axis) and distribution shift results (y-axis) with individual fine-tuned models (gray circles) and Model Soups~\cite{modelsoup}. 
    Note that \ours has much smaller ($35\times$ smaller) computational costs
than Model Soups, leveraging 71 various fine-tuned models as in the original paper.}
    \label{fig:appendix:lp_init}
\end{figure}

\section{Additional Experiments}
\label{suppl:additional_exp}
We present additional experimental studies to verify the effectiveness and applicability of \ours. 

\subsection{Experiments with LP initialization}
We conduct \ours with LP initialization and compare it with Model Soups that are initialized from LP. 
The results are in Fig.~\ref{fig:appendix:lp_init}.
In this experiment, we use the 16-epoch training setup with strong data augmentation for training \ours.
As shown in Fig.~\ref{fig:appendix:lp_init}, \ours outperforms the individual fine-tuned models\footnote{All the individual model checkpoints are from the official Model Soup repository.} (gray dots) on ImageNet accuracy.
\ours also demonstrates competitive performance against Model Soups considering WiSE-FT curves. 
Note that \ours is much more efficient ($35 \times$) than Model Soups, which utilize $71$ models in this experiment.

\begin{table}[t!]
\centering
\caption{\textbf{Complete results of Table~\ref{exp:vit_b_16}} with ObjectNet~\cite{objectnet} and ImageNet-ReaL~\cite{are_we_done}.}
\resizebox{0.8\linewidth}{!}{
\begin{tabular}{@{}l|cc|ccccc@{}}
\toprule
\multirow{2}{*}{Method} & \multicolumn{2}{c}{In-distribution} & \multicolumn{5}{c}{Distribution shifts}     \\ \cmidrule(l){2-8} 
                        & ImageNet          & IN-ReaL         & IN-V2 & IN-R & IN-A & IN-Sketch & ObjectNet \\ \midrule
Zero-shot               & 68.3              & 75.1            & 62.0  & \textbf{77.7} & {49.9} & 48.3      & {54.2}     \\
Vanilla FT              & 82.8              & 87.8            & 72.9  & 66.4 & 43.7 & 48.0      & 51.8     \\
Vanilla FT$^{*}$        & 83.7              & 87.8            & 73.5  & 67.6 & 40.0 & 48.6      & 50.1     \\
LP~\cite{lpft}                      & 79.7              & -               & 71.5  & 52.4 & 27.8 & 40.5      & -         \\
LP-FT~\cite{lpft}                   & 81.7              & -               & 71.6  & \underline{72.9} & 49.1 & 48.4      & -         \\
CAR-FT~\cite{car_ft}                  & 83.2              & -               & 73.0  & 71.3 & 43.7 & 49.5      & -         \\
FTP~\cite{tian2023fast}             & \underline{84.2}              & -               & 74.6  & 47.2 & 26.5 & 50.2      & -         \\
FLYP~\cite{flyp}                 &  82.6                & -                 &  73.0     &  71.4       &  48.1     &   49.6  & \textbf{58.7} \\
Lipsum-FT~\cite{lipsum-ft}                 &  83.3                & -                 &  73.6    & 75.9        &  49.9     & 51.4    & 54.4 \\
CaRot~\cite{carot}                 &  83.1                & -                 &   74.1    &  \textbf{77.7}      &   \textbf{51.6}    &  \textbf{52.7}   & \underline{56.6} \\
\rowcolor{Gray} \ours             & 84.1              & \underline{88.8}            & 74.8  & 71.8 & \underline{51.2} & \underline{51.8}      & {55.0}     \\
\rowcolor{Gray} \ourss             & \textbf{85.2}              & \textbf{89.1}            & \textbf{75.3}  & 68.7 & 45.0 & {51.3}      & 52.3     \\ 
\bottomrule
\end{tabular}
}
\label{tab:appendix:b16}
\end{table}

\subsection{Complete comparison results on CLIP ViT-B/16}
\label{suppl:vit-b16_complete}
In the main paper, we omit the results of ObjectNet~\cite{objectnet} on CLIP ViT-B/16 experiments since the comparison methods such as LP-FT~\cite{lpft}, FTP~\cite{tian2023fast} have not evaluated on ObjectNet benchmark. 
We here show the results with ObjectNet~\cite{objectnet} and ImageNet-ReaL~\cite{are_we_done} of CLIP ViT-B/16 in Table~\ref{tab:appendix:b16}.
We additionally compare \ours with recent fine-tuning methods including FLYP~\cite{flyp}, Lipsum-FT~\cite{lipsum-ft}, and CaRot~\cite{carot}
\ours consistently demonstrates its effectiveness with ObjectNet and ImageNet-ReaL as well.

\begin{table}[t]
\centering
\caption{
\textbf{Comparison against Model Soups~\cite{modelsoup} on CLIP ViT-B/16.}
\ours shows comparable performance with Model Soups. 
}
\label{tab:exp_vit_b16_soups}
\begin{tabular}{@{}lcc@{}}
\toprule
Method                                     & ImageNet  & Avg. shifts \\ \midrule
CLIP zero-shot Init.         & 68.3 & 58.4 \\
Vanilla FT                   & 82.8 & 56.6 \\
Vanilla FT$^\star$           & 83.7 & 55.9 \\
Uniform Model Soup           & \underline{84.4} & \textbf{62.7} \\
Greedy Model Soup            & 84.3  & 60.4 \\ \midrule
\rowcolor{Gray} \ours        & 84.1 & \underline{61.0} \\ 
\rowcolor{Gray} \ourss       & \textbf{85.2} & 58.5\\ 
\bottomrule
\end{tabular}
\label{exp:vit_b_16_model_soup_comp}
\end{table}

\subsection{\ours vs. Model Soups on CLIP ViT-B/16}
Table~\ref{exp:vit_b_16_model_soup_comp} shows the performance of \ours on the pretrained CLIP ViT-B/16 model. Since the original Model Soups paper~\cite{modelsoup} only provides CLIP ViT-B/32 models, we replicate Model Soups experiments on CLIP ViT-B/16. We fine-tuned 48 models from CLIP ViT-B/16 initialization following the standard grid hyper-parameter sweep (\ie, \textit{zero-shot} initialization setting).
\ours shows comparable performance against Model soups.
Note that Model Soups requires 24$\times$ more training cost than \ours.

\begin{table}[t!]
    \centering
    \caption{\textbf{\ours with different hyper-parameters on CLIP ViT-B/32}.    
    }
    \small 
    \begin{tabular}{@{}lcc@{}}
    \toprule
    Method &  ImageNet & Avg. shifts       \\ \midrule
    \ours                              &  79.89 & 50.99 \\
    \ours w/ different hyper-parameters & 79.75{$\pm$0.45} & 50.40{$\pm$0.84} \\    
    \bottomrule
    \end{tabular}
    \label{tab:appendix:diff_hp}
\end{table} 

\subsection{\ours with different hyper-parameters}
To verify the validity of \ours beyond the setup of the main paper (\ie, different random seeds with the same hyper-parameters), we conduct \ours with different hyper-parameters. 
In detail, when we fine-tune two models for \ours, we choose different hyper-parameter for each model (\eg, learning rate, data augmentation.). 
To ensure the basic assumption of \ours, we use the same batch size and training epochs. 
\ref{tab:appendix:diff_hp} shows the experimental results on CLIP ViT-B/32. 
We repeat 5 runs and report accuracy with standard deviation. 
\ours with different hyper-parameters shows comparable performance to the original one.

\begin{table}[t]
\centering
\small
\caption{\textbf{Performance comparison of merging units in \ours.} 
This table presents the overall performance of \ours using different merging units: entire weight merging, entire weight merging based on transformer block angle, layer-wise merging, and filter-wise merging. It highlights the effectiveness of each strategy in approaching the weight center and their impact on the model's performance.}
\begin{tabular}{lccc}
\toprule
\multicolumn{1}{c}{Merging Unit} &
  \multicolumn{2}{c}{Target} &
  \multirow{2}{*}{\begin{tabular}[c]{@{}c@{}}Avg.\\ Shifts\end{tabular}} \\ \cmidrule{2-3}
                                                            & IN    & IN-ReaL &       \\ \midrule
Entire weights                                              & 79.69 & 85.39   & 46.40  \\
Entire weights (rep.\ blocks only)&
  79.64 &
  85.38 &
  48.28 \\
\rowcolor{Gray} Layer-wise (ours) & \textbf{80.12} & \underline{85.65}   & \textbf{48.84} \\
Filter-wise                                                 & \underline{80.10} & \textbf{85.67}   & \underline{48.72} \\
\bottomrule
\end{tabular}
\label{tab:merge_unit}
\end{table}
\subsection{Ablation study on merging unit}
\label{para:merging_unit}
We investigate the efficacy of different merging units within our method, \ours. 
Our default approach employs layer-wise merging, but alternatives include merging based on the angle between 1) entire weights, 2) weights of the entire repetitive transformer blocks following \cite{modelsoup}, or 3) using a filter-wise approach as discussed in \S\ref{suppl:filter_wise}. 
The results of these ablations are summarized in Table~\ref{tab:merge_unit}, where we assess the overall performance based on the chosen merging unit.

Our analysis reveals that the accuracy of noise distribution estimation is critical in approaching the weight center. 
When assuming weight noise across the entire model, our method does not approximate the weight center as effectively as it does with layer-wise merging, leading to suboptimal overall performance. 
Similarly, the merging performance based on the angle of transformer blocks was insufficient.
Conversely, while filter-wise noise demonstrates a larger standard deviation in angle, as depicted in Fig.~\ref{fig:angle_filter}, this increased variance results in a more significant error in Gaussian distribution approximation. 
Consequently, the overall performance under filter-wise merging is slightly inferior to layer-wise one.

These findings underscore the importance of accurately modeling noise distribution in enhancing the performance of \ours. 
As our understanding and ability to model this noise distribution improve, we anticipate further increases in the efficacy and robustness of our approach.

\begin{figure*}[t]
    \vspace{-2.5em}
    \centering
    \begin{subfigure}[t]{\linewidth}
    \centering
        \includegraphics[width=0.98\linewidth]{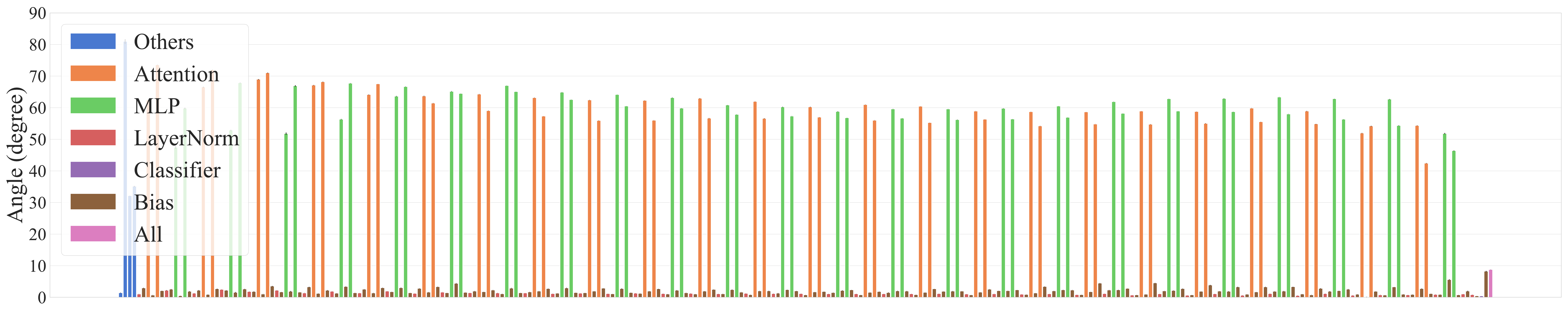}
        \includegraphics[width=0.98\linewidth]{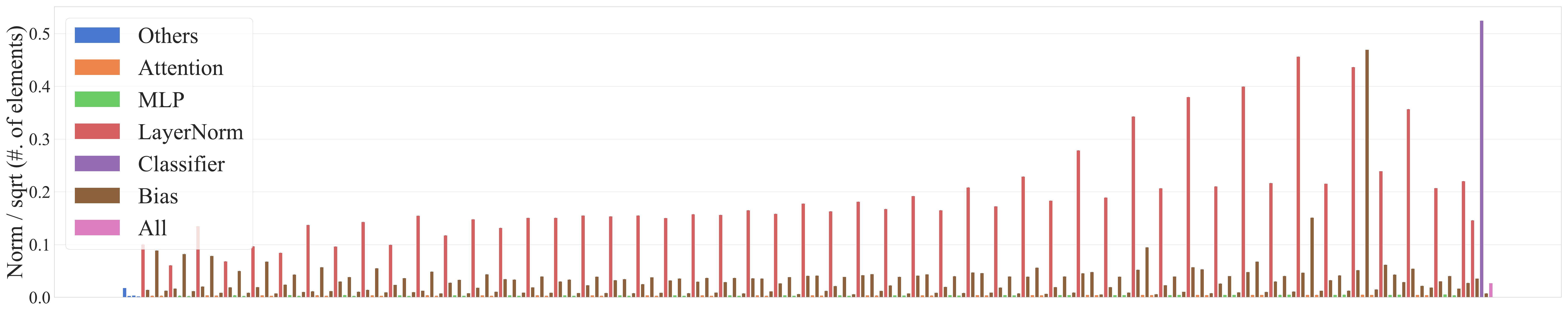}
        \caption{CLIP ViT-L/14}
    \end{subfigure}
    
    \begin{subfigure}[t]{\linewidth}
    \centering
        \includegraphics[width=0.98\linewidth]{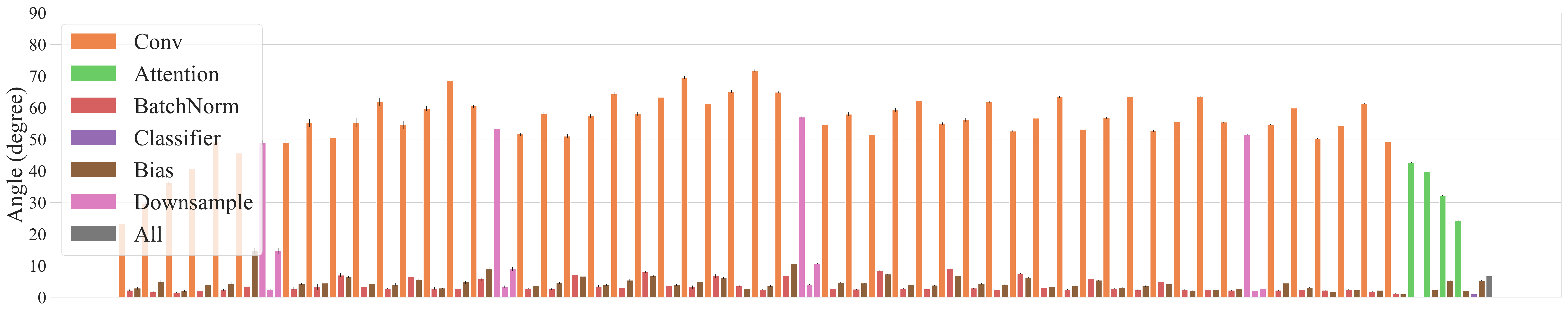}
        \includegraphics[width=0.98\linewidth]{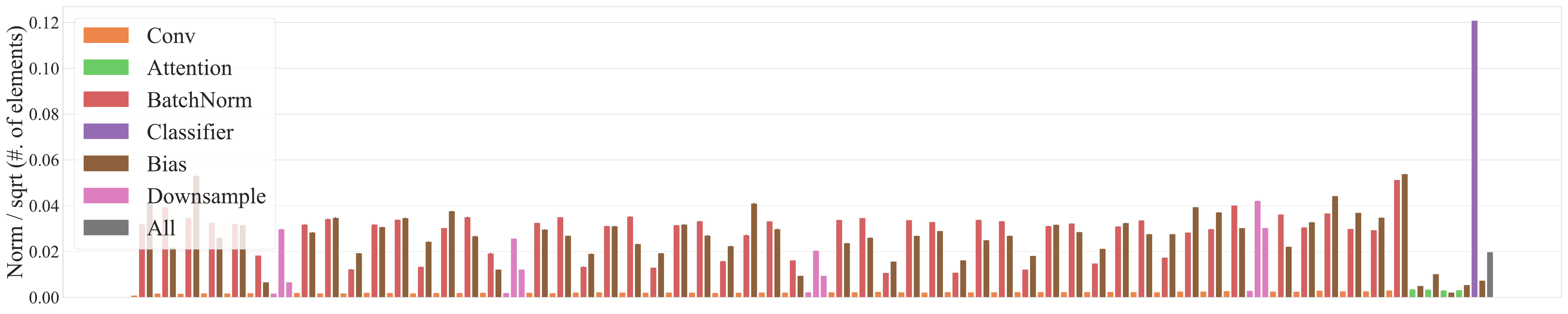}
        \caption{CLIP ResNet50}
    \end{subfigure}
    
    \begin{subfigure}[t]{\linewidth}
    \centering
        \includegraphics[width=0.98\linewidth]{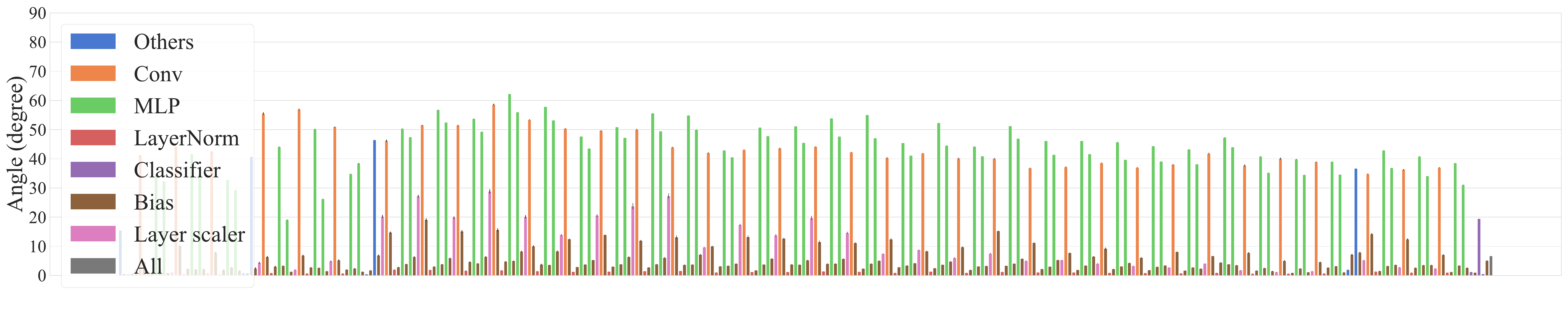}
        \includegraphics[width=0.98\linewidth]{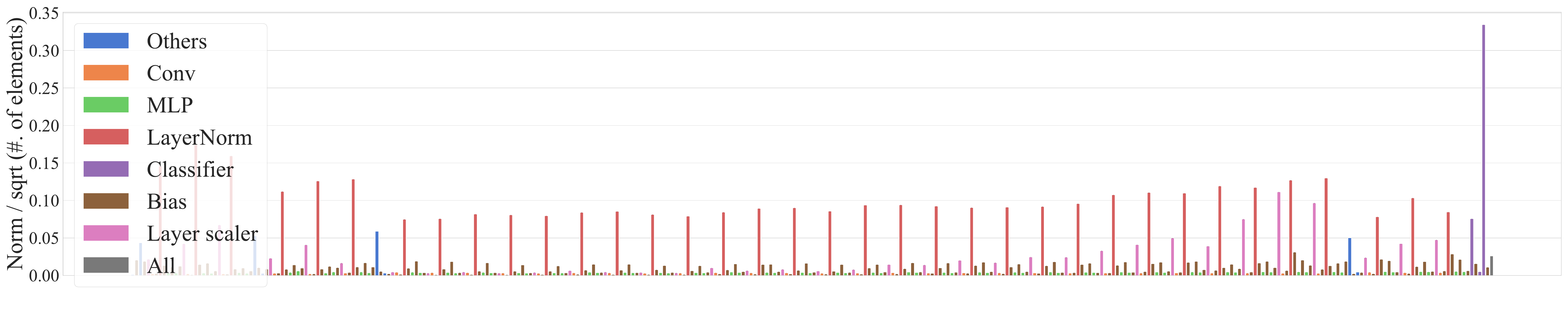}
        \caption{OpenCLIP ConvNeXt}
    \end{subfigure}
    
        \vspace{-.75em}
        \caption{\textbf{Layer-wise angle and norm across different model architectures.}
    The angle and norm for CLIP ViT-L/14, CLIP ResNet50, and OpenCLIP ConvNeXt are displayed from top to bottom. These metrics demonstrate consistency regardless of the model type from left (first layer) to right (last layer). It is important to note that we also depict the error bars for each layer in all figures, but they are not visible in most layers due to the small standard deviation.}
        \label{fig:angle_arch}
        \vspace{-.5em}
    \end{figure*}

\begin{figure*}[t]
\centering
    \begin{subfigure}[t]{\linewidth}
    \centering
        \includegraphics[width=0.98\linewidth]{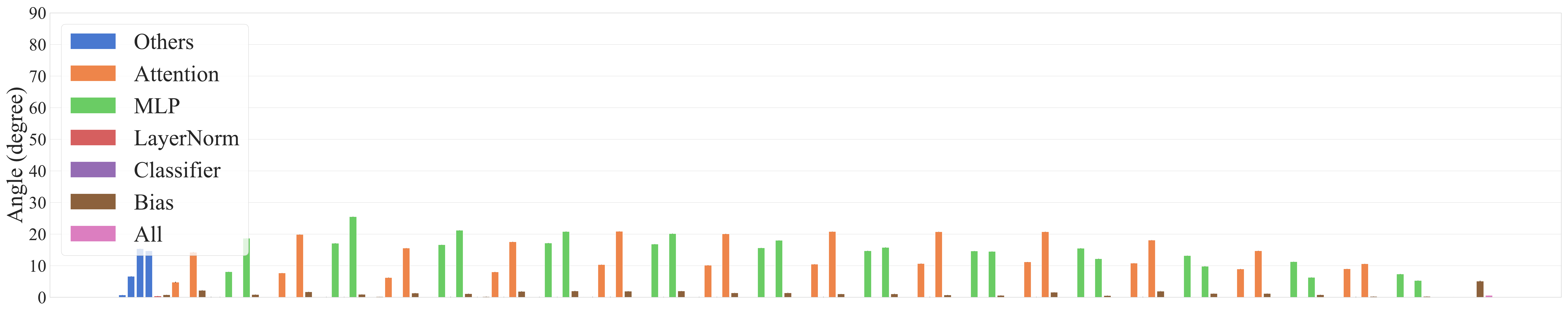}
        \includegraphics[width=0.98\linewidth]{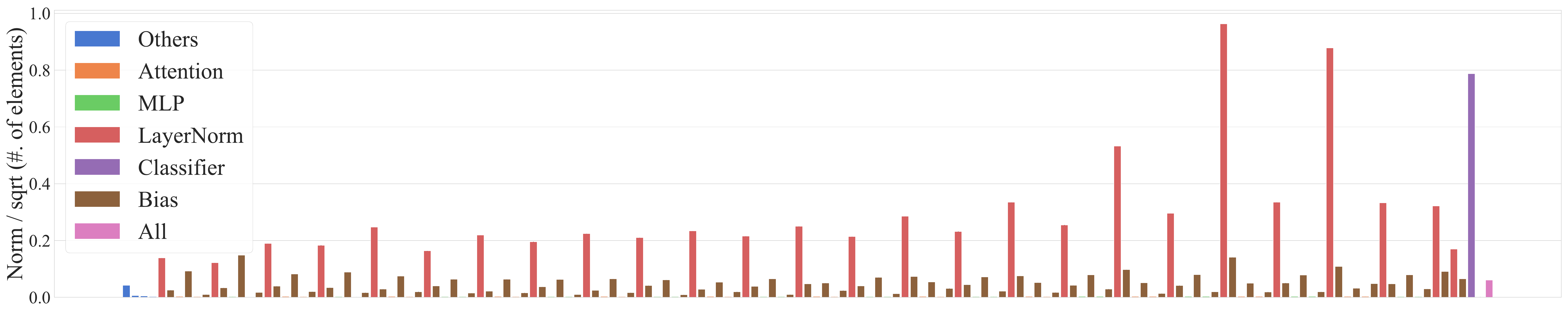}
        \caption{SGD optimizer}
    \end{subfigure}
    \begin{subfigure}[t]{\linewidth}
    \centering
        \includegraphics[width=0.98\linewidth]{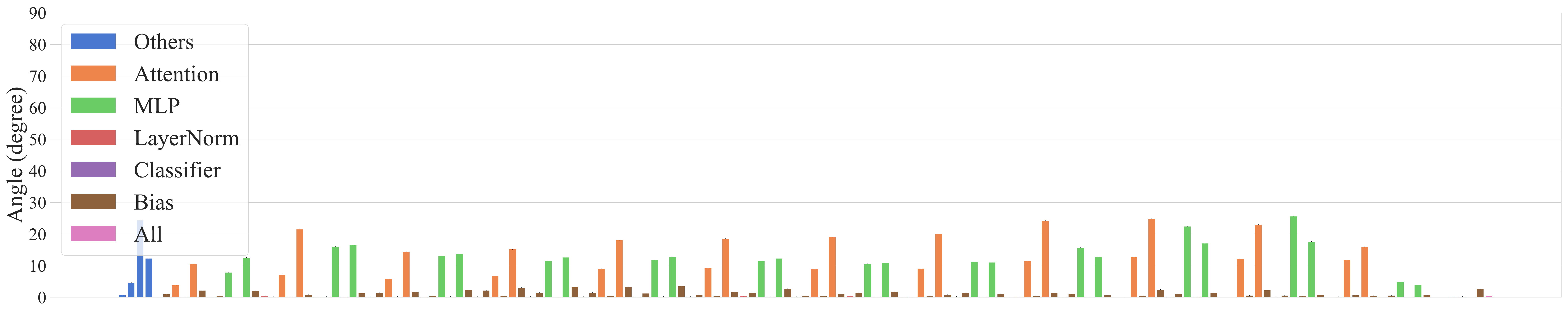}
        \includegraphics[width=0.98\linewidth]{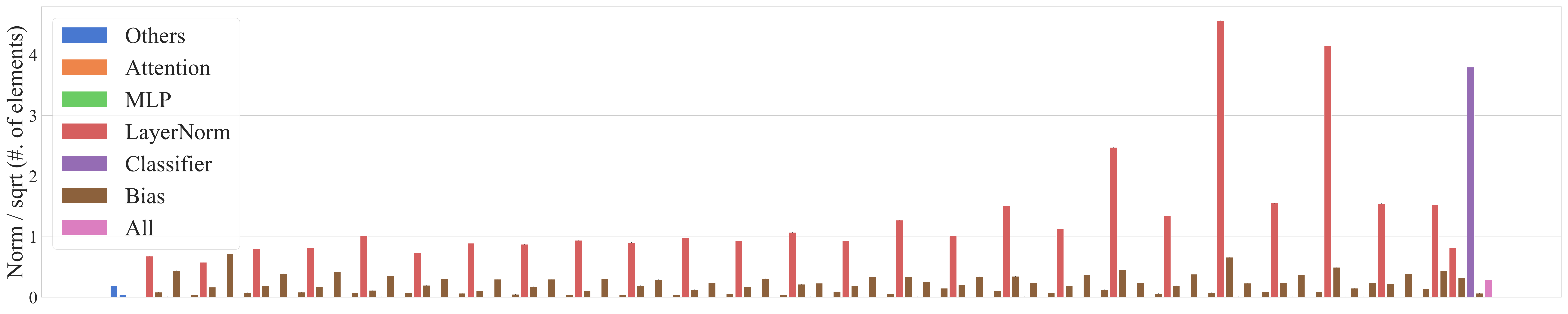}
        \caption{SGD optimizer with momentum}
    \end{subfigure}
    \vspace{-.75em}
    \caption{\textbf{Layer-wise angle and norm across different optimizers.}
Displayed from top to bottom are the angle and norm for models trained with SGD and SGD with momentum, respectively. These metrics demonstrate consistency regardless of the optimization strategy from left (first layer) to right (last layer).}
    \label{fig:angle_optim}
    \vspace{-.5em}
\end{figure*}

\begin{figure*}[t]
\vspace{-2.5em}
\centering
    \begin{subfigure}[t]{\linewidth}
    \centering
        \includegraphics[width=0.98\linewidth]{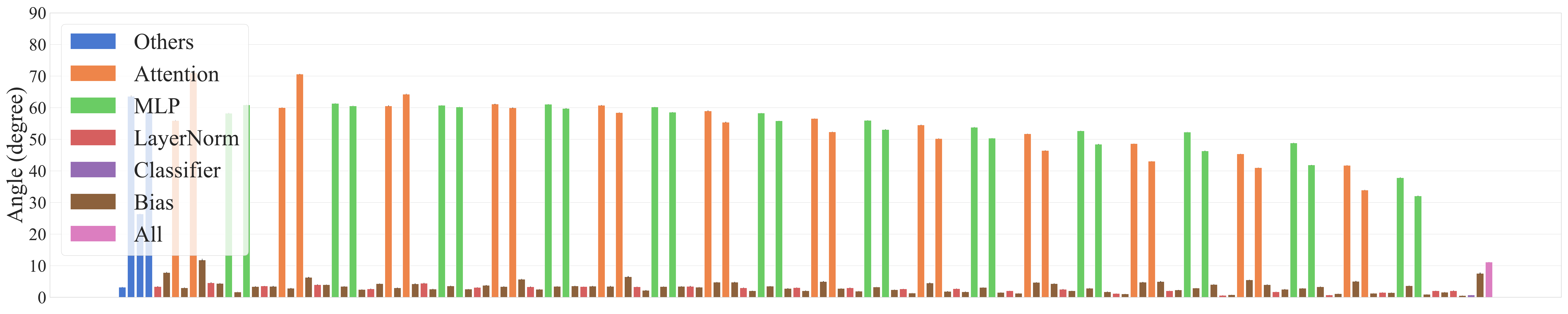}
        \includegraphics[width=0.98\linewidth]{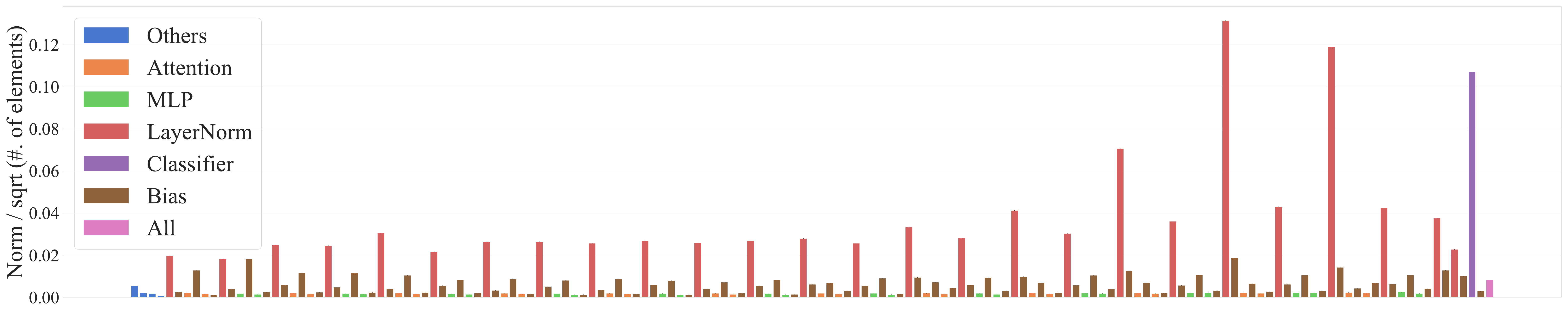}
        \caption{Vanilla model (10 epochs + no augmentation)}
    \end{subfigure}
    \begin{subfigure}[t]{\linewidth}
    \centering
        \includegraphics[width=0.98\linewidth]{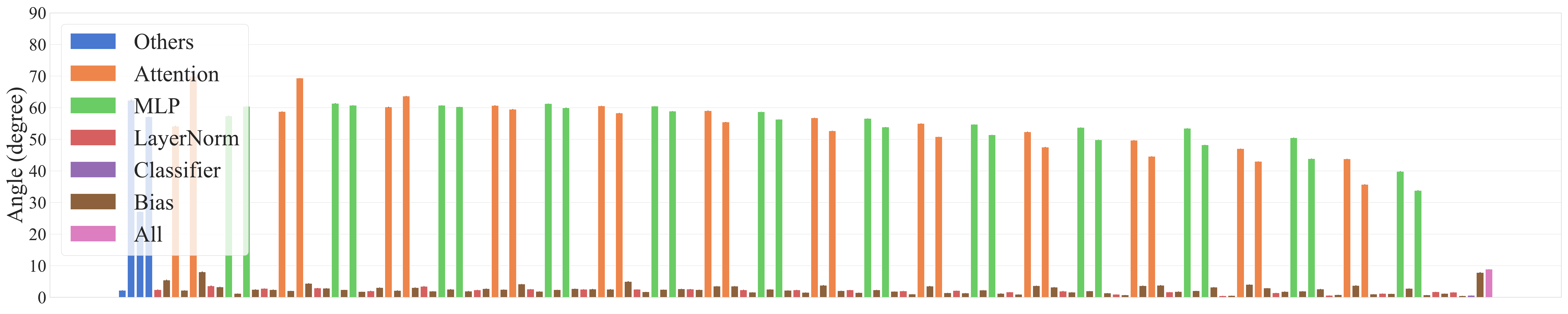}
        \includegraphics[width=0.98\linewidth]{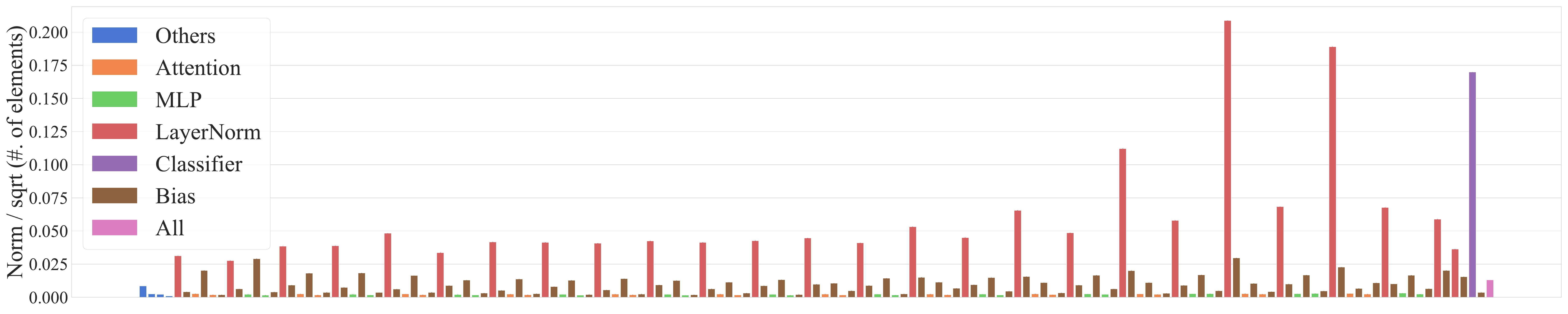}
        \caption{+ longer epochs (16 epochs)}
    \end{subfigure}
    \begin{subfigure}[t]{\linewidth}
    \centering
        \includegraphics[width=0.98\linewidth]{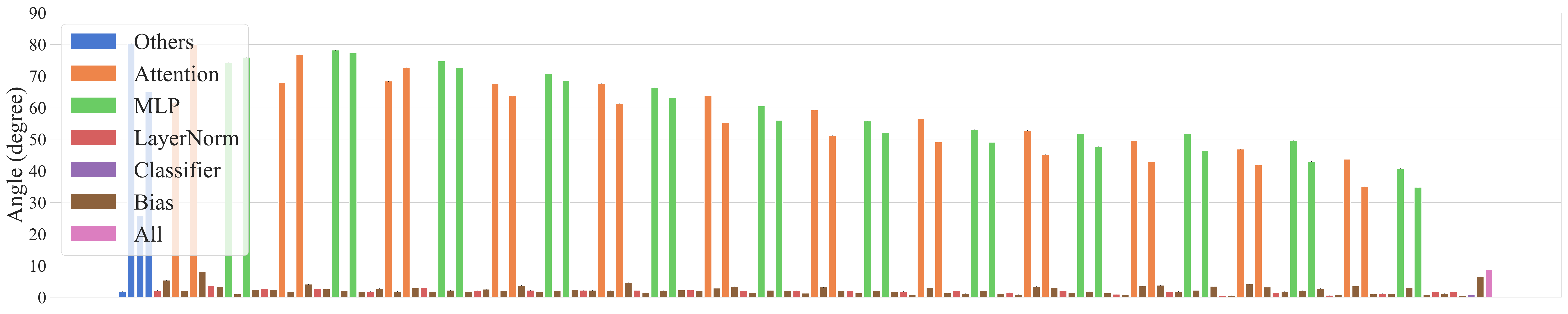}
        \includegraphics[width=0.98\linewidth]{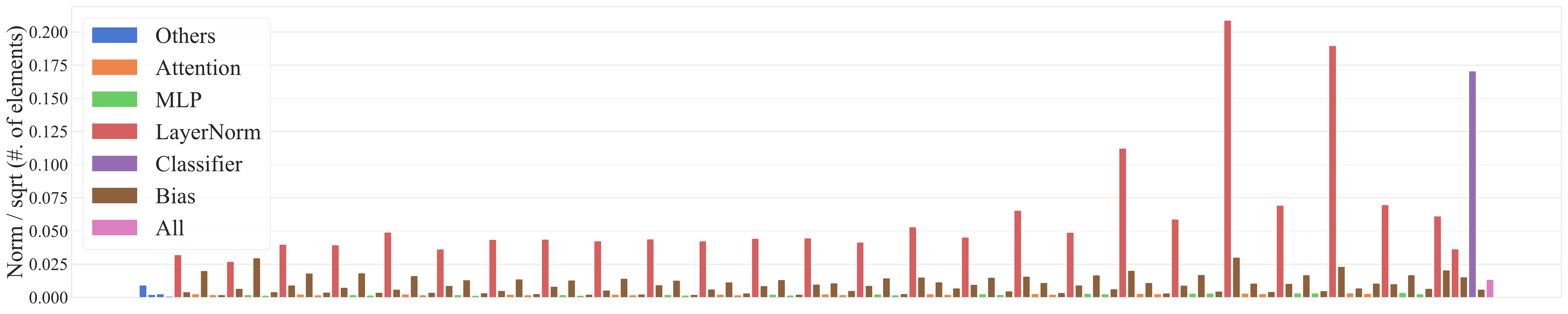}
        \caption{+ RRC}
    \end{subfigure}
    \vspace{-.75em}
    \caption{\textbf{Layer-wise angle and norm across different augmentations.}
    Displayed from top to bottom are the angle and norm for the vanilla model (10 epochs + no augmentation), +longer epochs (16 epochs), and +RRC. Each augmentation is applied incrementally. These metrics demonstrate consistency regardless of the augmentations from left (first layer) to right (last layer).}
    \label{fig:angle_aug}
    \vspace{-.5em}
\end{figure*}

\begin{figure*}[t]
\centering
    \includegraphics[width=0.98\linewidth]{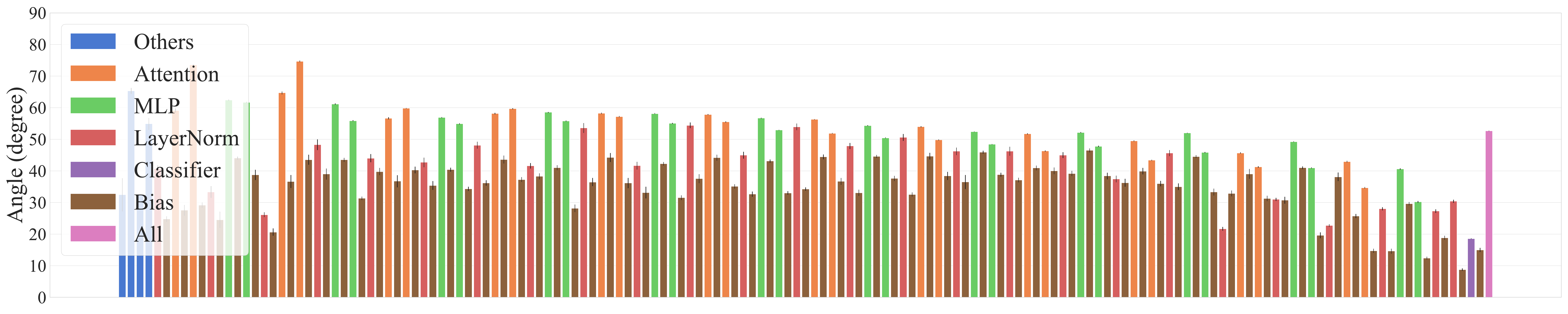}
    \includegraphics[width=0.98\linewidth]{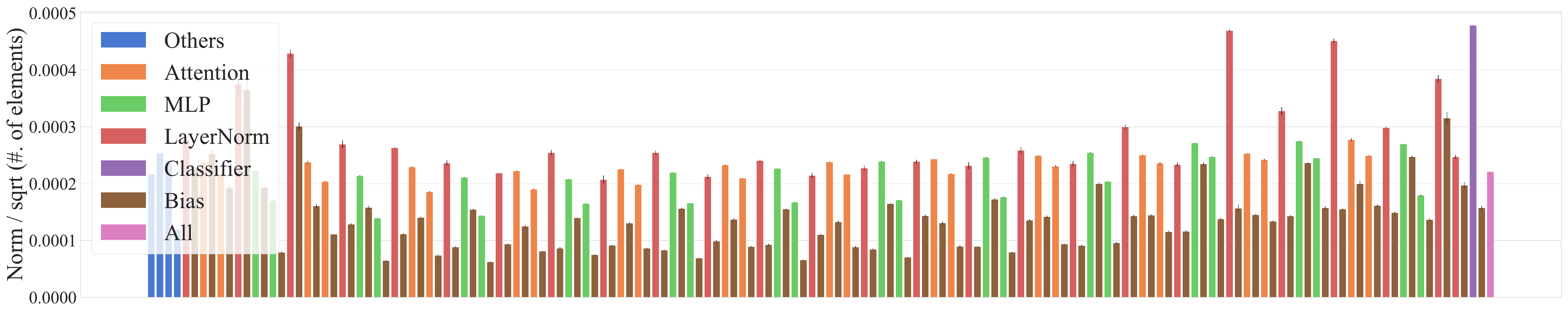}
    \vspace{-.75em}
    \caption{\textbf{Layer-wise angle and norm across different datasets.}
The angle and norm for models trained on different datasets, including CIFAR~\cite{cifar} are displayed from top to bottom. These metrics demonstrate consistency regardless of the dataset type from left (first layer) to right (last layer).}
    \label{fig:angle_dataset}
    \vspace{-.5em}
\end{figure*}

\begin{figure*}[t]
\centering
    \includegraphics[width=0.98\linewidth]{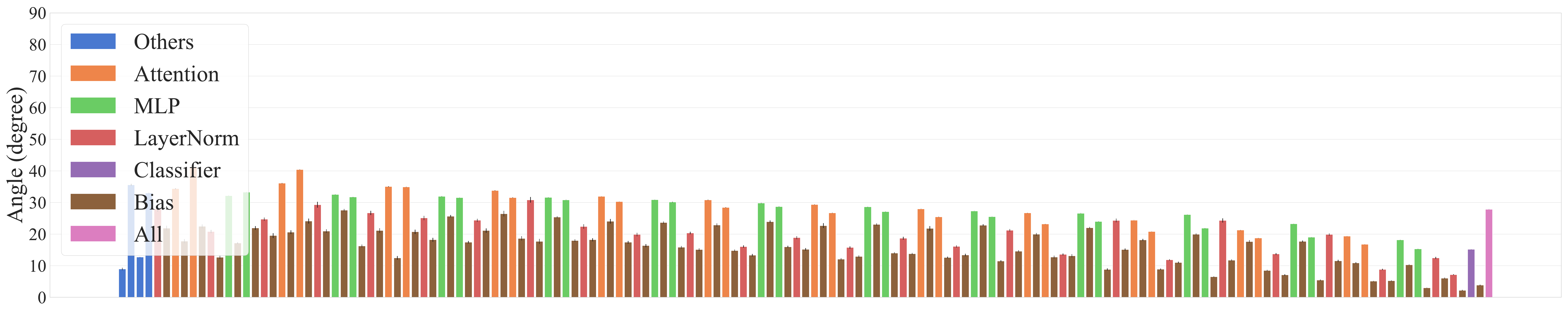}
    \includegraphics[width=0.98\linewidth]{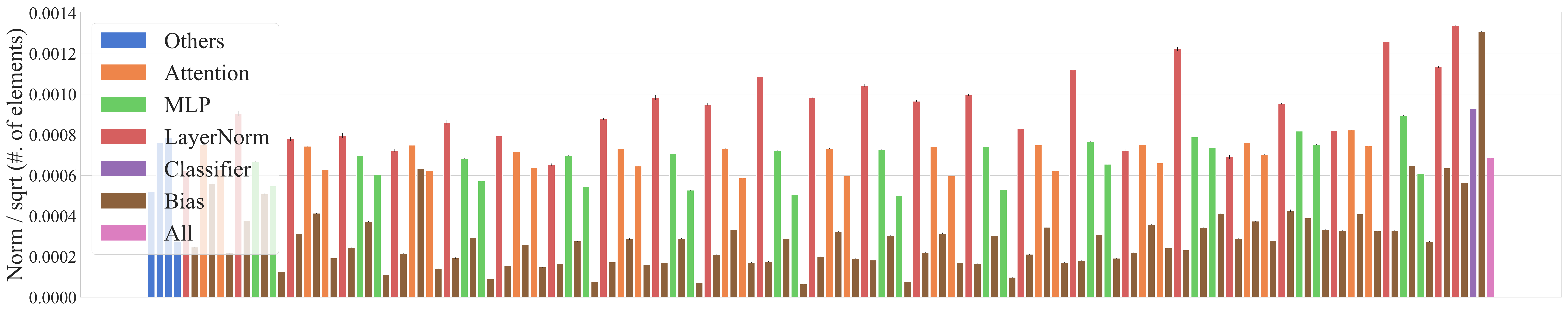}
    \vspace{-.75em}
    \caption{\textbf{Layer-wise angle and norm across different classifier initializations.}
The angle and norm for models trained with differently initialized networks following the LP-FT~\cite{lpft} method are displayed from top to bottom. These metrics demonstrate consistency regardless of the initialization method from left (first layer) to right (last layer).}
    \label{fig:angle_init}
    \vspace{-.5em}
\end{figure*}

\begin{figure*}[t]
\centering
    \includegraphics[width=0.98\linewidth]{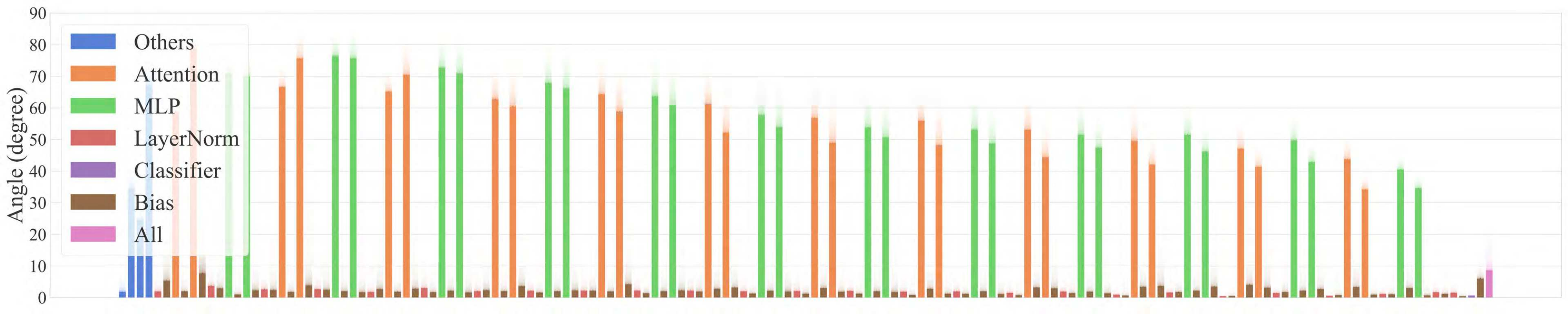}
    \vspace{-.75em}
    \caption{\textbf{Layer-wise angle during training.}
Displayed are the overlapped angles across models trained with different random seeds at each timestamp. Even during training, the angle remains highly consistent, decreasing as training progresses.}
    \label{fig:angle_during}
    \vspace{-.5em}
\end{figure*}

\begin{figure*}[t]
\centering
    \begin{subfigure}[t]{\linewidth}
        \centering    
        \includegraphics[width=0.98\linewidth]{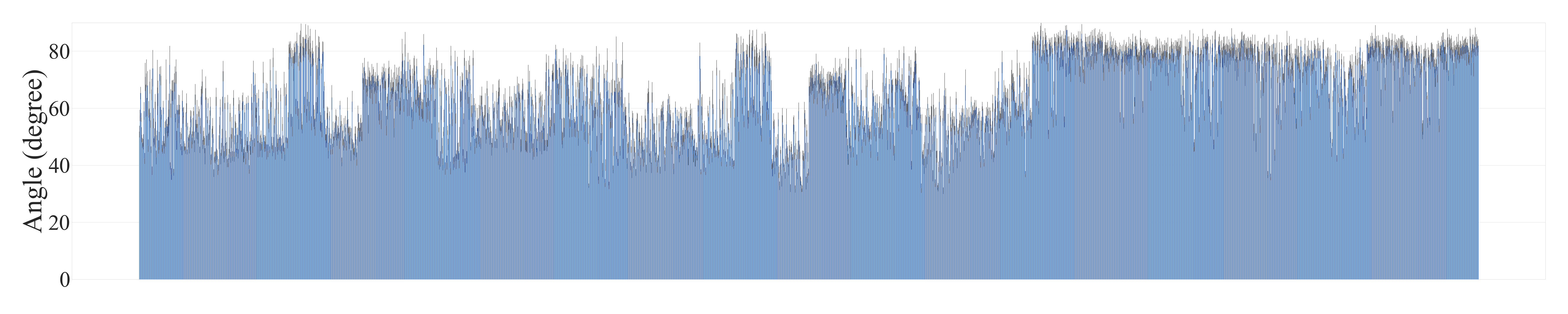}
        \vspace{-.5em}
        \caption{Filter-wise angle between attention weights in the first transformer block of ViT-B/32}
        \vspace{-.25em}
    \end{subfigure}
    \begin{subfigure}[t]{\linewidth}
        \centering    
        \includegraphics[width=0.98\linewidth]{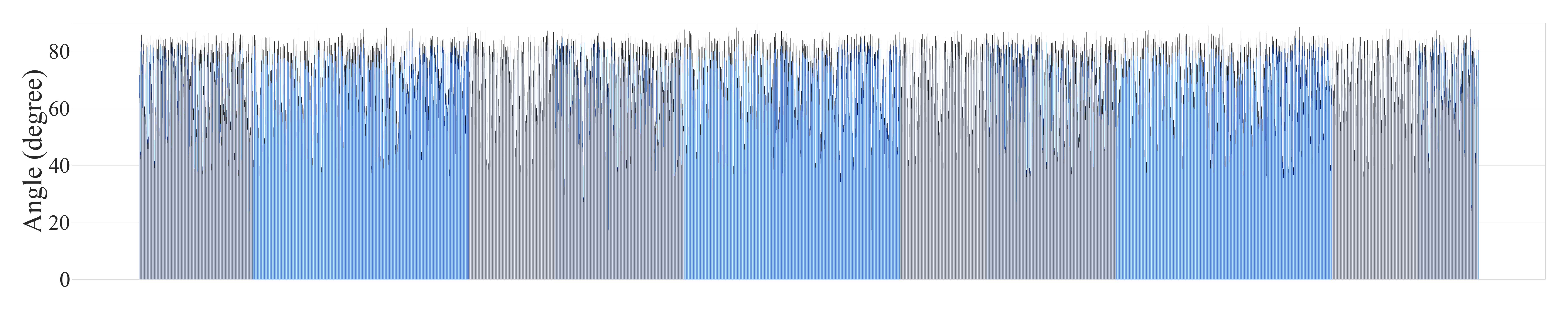}
        \vspace{-.5em}
        \caption{Filter-wise angle between MLP weights in the first transformer block of ViT-B/32}
        \vspace{-.25em}
    \end{subfigure}
    \begin{subfigure}[t]{\linewidth}
        \centering    
        \includegraphics[width=0.98\linewidth]{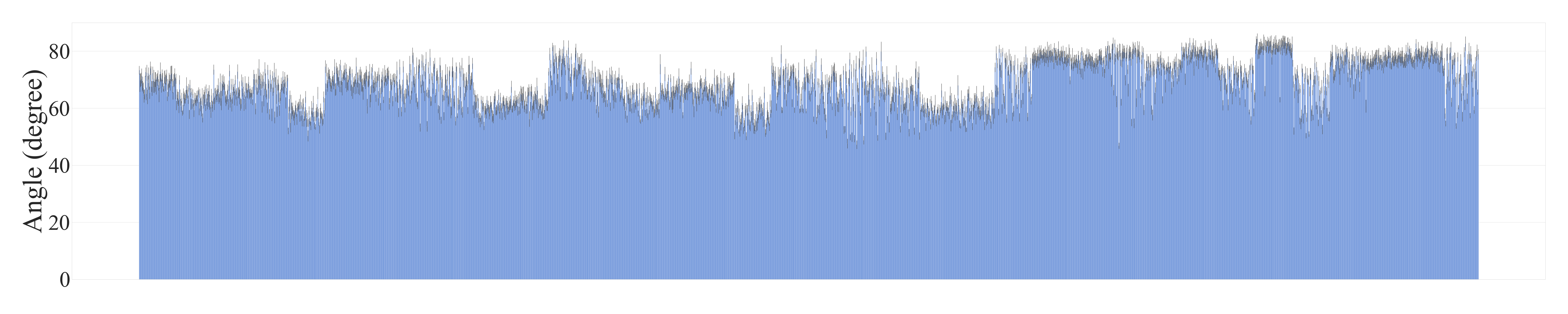}
        \vspace{-.5em}
        \caption{Filter-wise angle between attention weights in the second transformer block of ViT-B/32}
        \vspace{-.25em}
    \end{subfigure}
    \caption{\textbf{Filter-wise angle for attention and MLP layers in ViT-B/32.} We display filter-wise angles for each layer. 
Each bar represents each row (\ie, filter) in the given layer.
Interestingly, the angles between the filters of the fine-tuned weights exhibit similar values, while the standard deviation between each filter is notably larger than that of the angle between each layer. 
Due to the large number of layers, only representative layers are selected for display.}
    \label{fig:angle_filter}
    \vspace{-.5em}
\end{figure*}

\begin{figure*}[t]
\centering
    \includegraphics[width=0.98\linewidth]{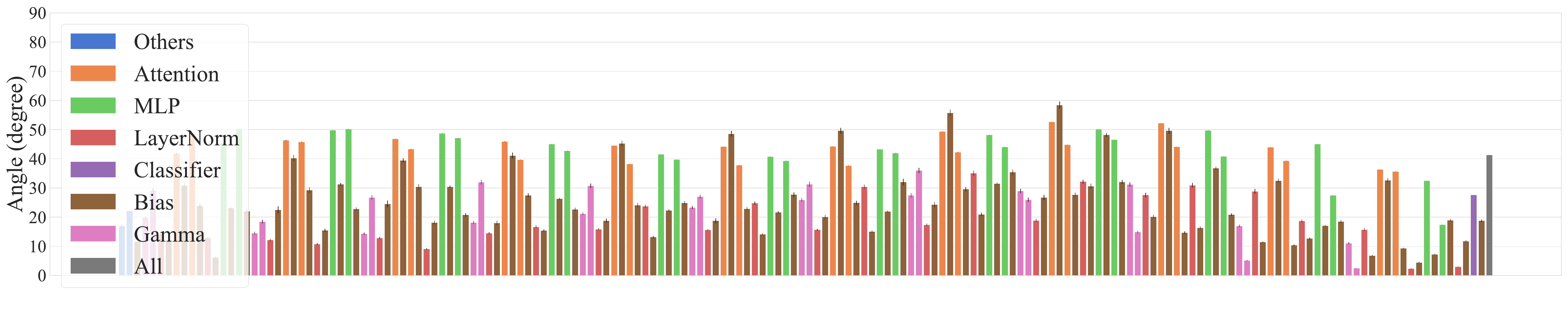}
    \includegraphics[width=0.98\linewidth]{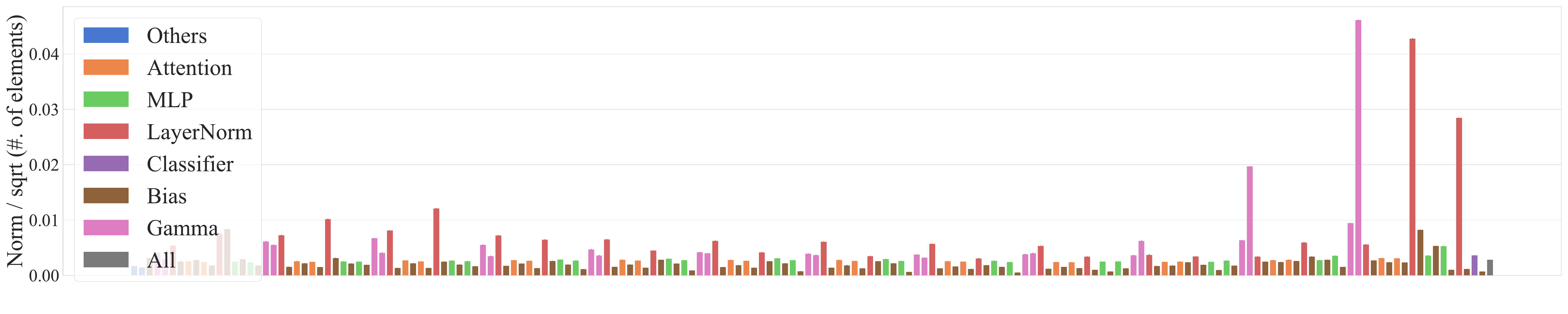}
    \vspace{-.75em}
    \caption{\textbf{Layer-wise angle and norm for DeiT.}
The angle and norm for DeiT-base models are displayed, each trained with different random seeds. These models are initially pre-trained on ImageNet-21K~\cite{imagenet} and then fine-tuned on ImageNet-1K. The consistency observed in the metrics is maintained even in the DeiT training setting.}
    \label{fig:angle_deit}
    \vspace{-.5em}
\end{figure*}

\end{document}